\documentclass[11pt]{article}

\usepackage[preprint]{acl}

\usepackage{times}
\usepackage{latexsym}

\usepackage{multirow}
\usepackage{makecell}

\usepackage[T1]{fontenc}
\usepackage[utf8]{inputenc}

\usepackage{microtype}
\usepackage{inconsolata}
\usepackage{graphicx}

\usepackage{multirow}
\usepackage{booktabs}
\usepackage{float}
\usepackage{subcaption}
\usepackage{amsmath}
\usepackage{amssymb}
\usepackage{mathtools}
\usepackage{amsthm}
\usepackage[capitalize,noabbrev]{cleveref}
\usepackage[textsize=tiny]{todonotes}

\theoremstyle{plain}

\theoremstyle{definition}

\theoremstyle{remark}

\title{From Unfamiliar to Familiar: Detecting Pre-training Data \\via Gradient Deviations in Large Language Models}

\author{
  \textbf{Ruiqi Zhang\textsuperscript{1,2,*}},
  \textbf{Lingxiang Wang\textsuperscript{1,2,*}},
  \textbf{Hainan Zhang\textsuperscript{1,2}},
  \textbf{Zhiming Zheng\textsuperscript{1,2}},
  \textbf{Yanyan Lan\textsuperscript{1,2}}
\\
\\
  \textsuperscript{1} Beijing Advanced Innovation Center for Future Blockchain and Privacy Computing, Beihang University \\
  \textsuperscript{2} School of Artificial Intelligence, Beihang University \\
  \textsuperscript{3} Institute for AI Industry Research (AIR), Tsinghua University \\
\\
  \small{
    \textbf{Correspondence:} \href{mailto:zhanghainan@buaa.edu.cn}{zhanghainan@buaa.edu.cn}
  }
}

\begin{document}
\maketitle

\begin{abstract}
Pre-training data detection for LLMs is essential for addressing copyright concerns and mitigating benchmark contamination. Existing methods mainly focus on the likelihood-based statistical features or heuristic signals before and after fine-tuning, but the former are susceptible to word frequency bias in corpora, and the latter strongly depend on the similarity of fine-tuning data. From an optimization perspective, we observe that during training, samples transition from unfamiliar to familiar in a manner reflected by systematic differences in gradient behavior. Familiar samples exhibit smaller update magnitudes, distinct update locations in model components, and more sharply activated neurons. Based on this insight, we propose GDS, a method that identifies pre-training data by probing Gradient Deviation Scores of target samples. Specifically, we first represent each sample using gradient profiles that capture the magnitude, location, and concentration of parameter updates across FFN and Attention modules, revealing consistent distinctions between member and non-member data. These features are then fed into a lightweight classifier to perform binary membership inference. Experiments on five public datasets show that GDS achieves state-of-the-art performance with significantly improved cross-dataset transferability over strong baselines. Further interpretability analyses reveal differences in gradient distributions, and the semi-supervised results offer a practical way to detect pre-training data.
\end{abstract}


\section{Introduction}

Large language models (LLMs) performance scales with the size and quality of pre-training data~\cite{kaplan2020scaling}. As pre-training corpora expand to trillions of tokens and become increasingly proprietary and non-transparent, they raise significant concerns, such as unauthorized copyright, biased or harmful content, and contamination of evaluation benchmarks~\cite{balloccu2024leak, grynbaum2023times}. These challenges motivate a specialized task of membership inference attacks, named pre-training data detection: determining whether a given text sample was included in a model’s pre-training corpus~\cite{carlini2021extracting}.

\begin{figure}[!t]
    \centering
    \includegraphics[width=0.38\textwidth, height=0.24\textheight]{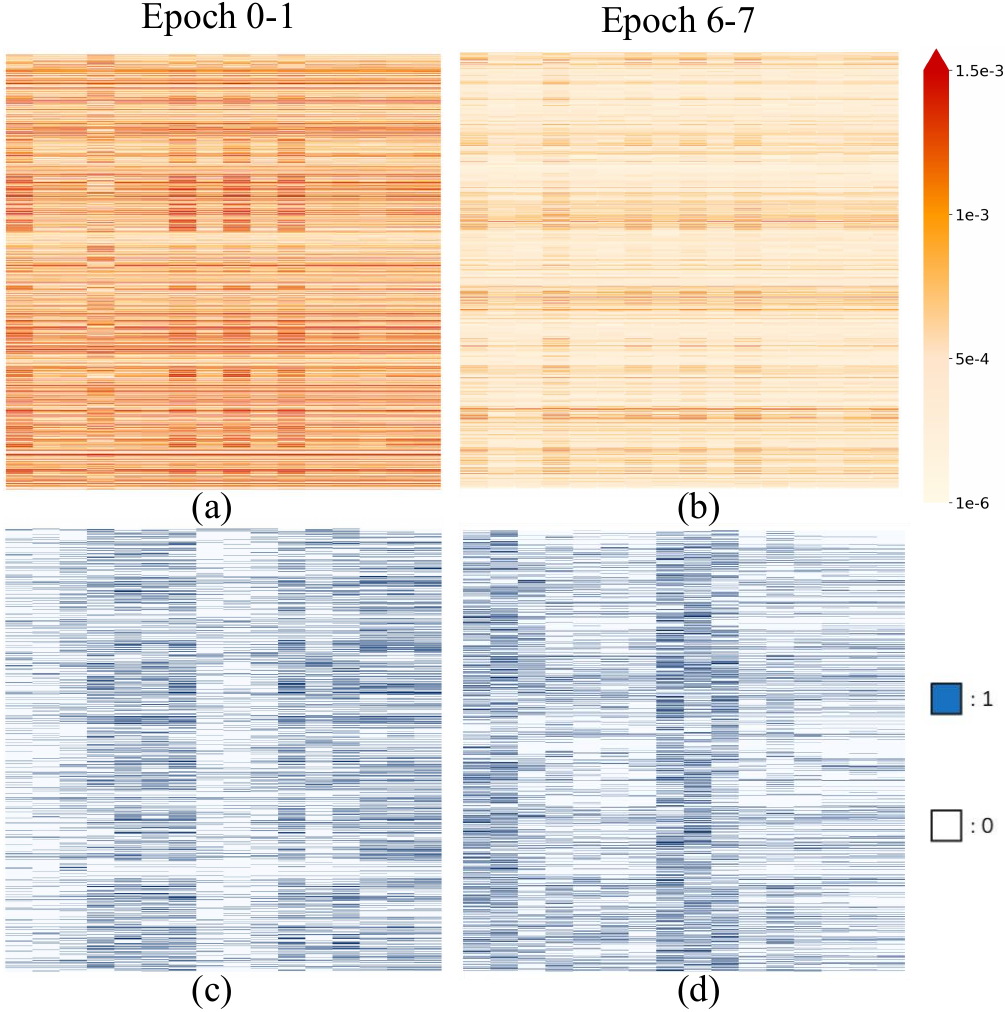}
    \caption{Heat maps of gradient matrices from unfamiliar model(a, c) and familiar model(b, d). Panels a and b display the update magnitude of the FFN module in layer 30, while panels c and d illustrate the update positions, with white for sparse and blue for core updates.}
    \label{fig:motivation1}
    \vspace{-12pt}
\end{figure}

Recent studies~\cite{carlini2021extracting,li2025estimating,shi2024detecting,zhang2025min} primarily use unsupervised likelihood-based statistical features to detect pre-training samples. For example, the Min-K method~\cite{shi2024detecting,zhang2025min} examines the k\% of tokens with the lowest predicted probabilities, assuming non-member texts contain more low-likelihood tokens. However, these approaches are susceptible to word frequency bias in pre-training corpora, specifically for rare words or short texts scenarios. To improve detection accuracy, other researchers~\cite{zhangfine,choicontaminated} propose some supervised methods to analyze heuristic signals before and after fine-tuning, leveraging the fact that fine-tuning impacts member and non-member samples differently. For example, KDS~\cite{choicontaminated} shows that non-member datasets exhibit much larger embedding changes, while FSD~\cite{zhangfine} finds that their loss reductions are also significantly greater. Nevertheless, these supervised methods rely on the strong assumption that fine-tuning data closely match the target samples' distribution, requiring additional training on similar non-member data and thereby limiting cross-dataset generalization. Therefore, how to achieve both high accuracy and strong generalization performance remains an open challenge for pre-training data detection task.

\begin{figure*}[!t]
    \centering
    \begin{subfigure}{0.22\textwidth} 
        \centering
        \includegraphics[width=\textwidth, height=0.93\textwidth]{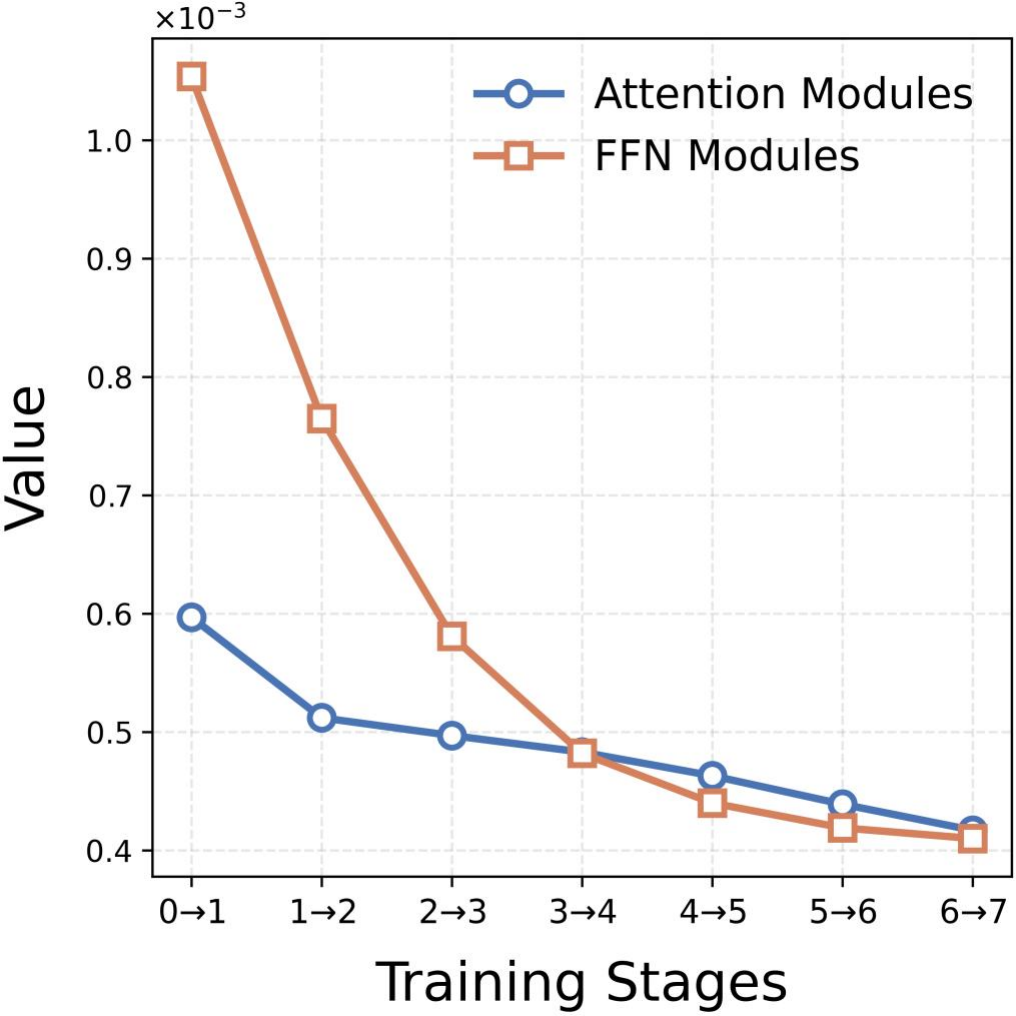}
        \caption{$\Delta\Theta_t$}
        \label{subfig:mean_trend}
    \end{subfigure}
    \hfill
    \begin{subfigure}{0.22\textwidth}
        \centering
        \includegraphics[width=\textwidth, height=0.93\textwidth]{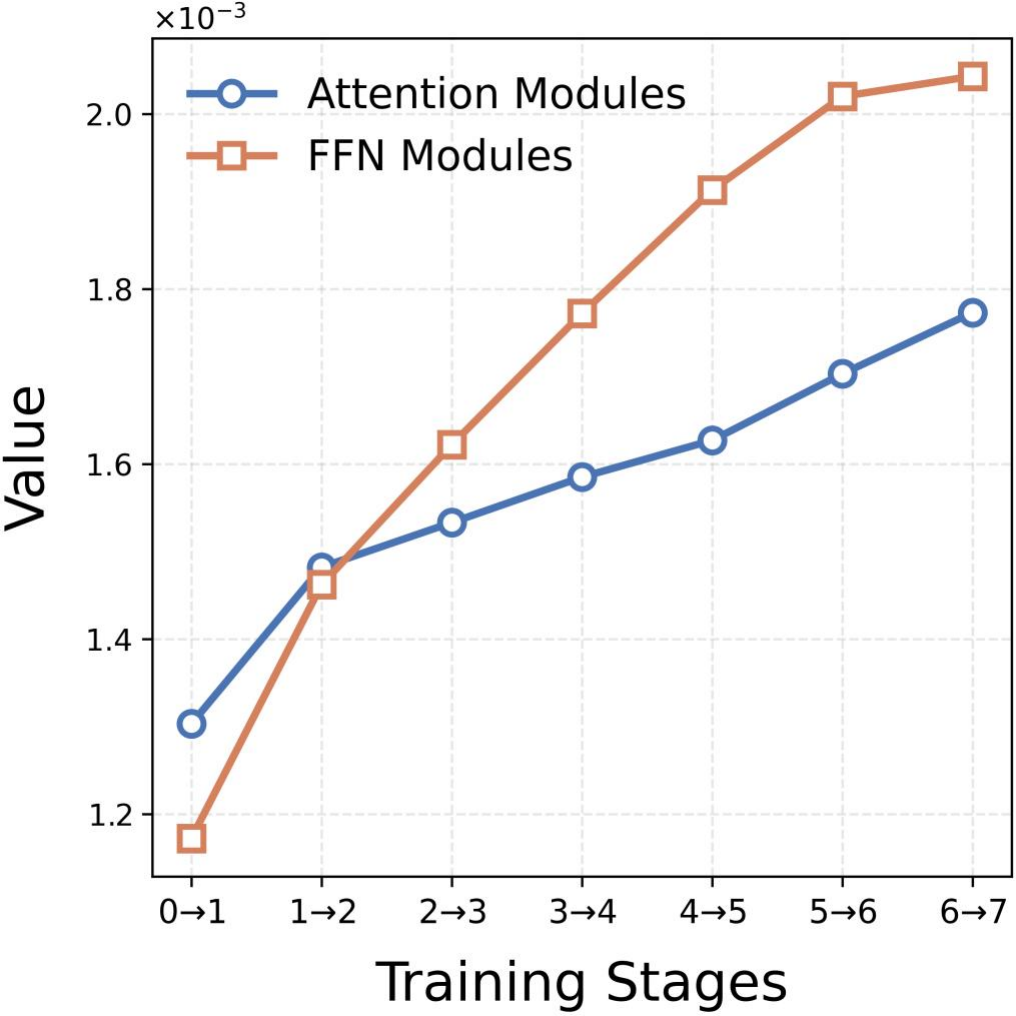}
        \caption{$S_t$}
        \label{subfig:sparsity_trend}
    \end{subfigure}
    \hfill
    \begin{subfigure}{0.22\textwidth}
        \centering
        \includegraphics[width=\textwidth, height=0.93\textwidth]{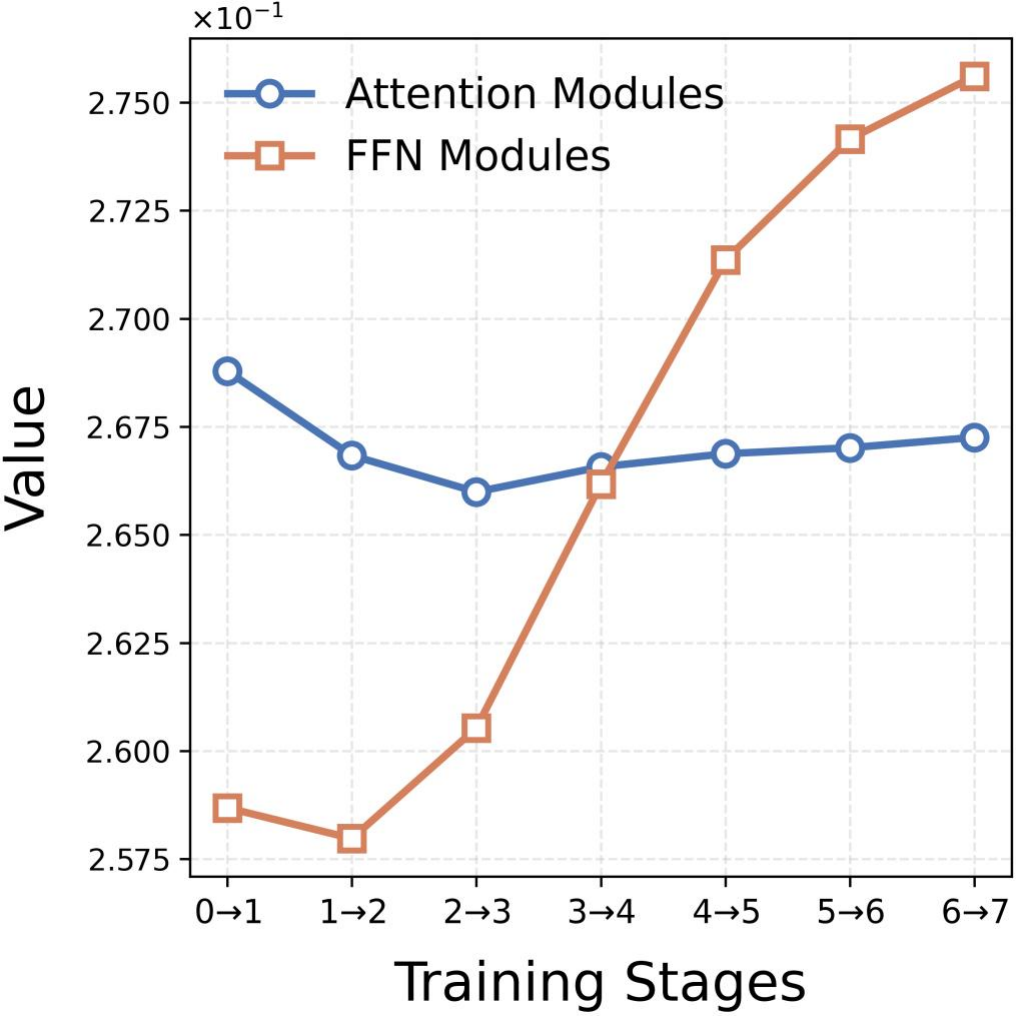}
        \caption{$T_{10,t}$}
        \label{subfig:eccentricity_trend}
    \end{subfigure}
    \hfill
    \begin{subfigure}{0.22\textwidth}
        \centering
        \includegraphics[width=\textwidth, height=0.93\textwidth]{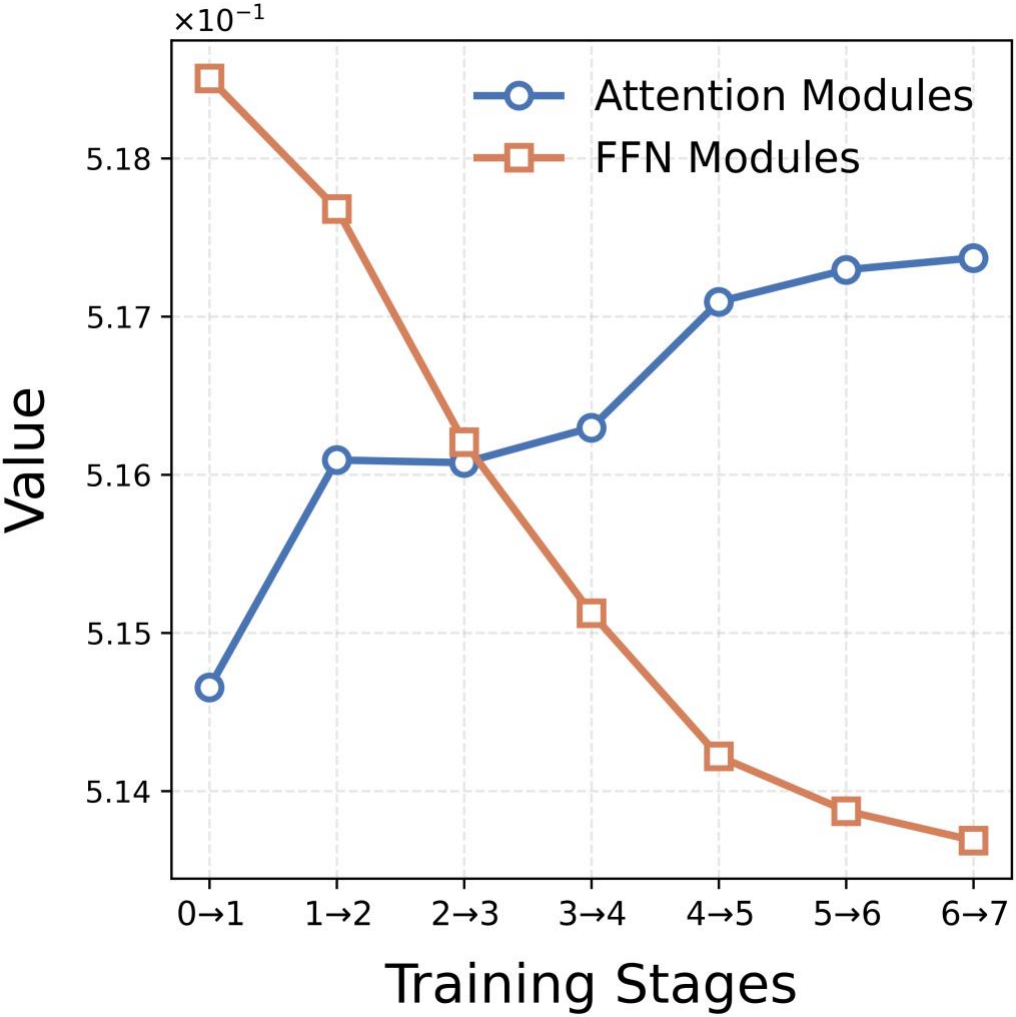} 
        \caption{$E_t$}
        \label{subfig:other_trend}
    \end{subfigure}
    \caption{Dynamic curves of four parameter update characteristics defined in the Motivation section. $i \to j$ denotes the epoch interval. Ordinate shows values of predefined characteristics, with orange and blue representing Attention and FFN modules.}
    \label{fig:motivation_four_features}
    \vspace{-12pt}
\end{figure*}

Optimization theory~\cite{ruder2016overview, frankleearly, wang2024q} suggests that training samples undergo a shift in gradient behavior as they move from unfamiliar to familiar. That is, pre-training member data and non-member data induce distinct gradients in target LLMs. Motivated by this, we compare LLMs gradient behaviors on familiar and unfamiliar data\footnote{Due to unavailable publish time, original LLaMA-7B is treated as unfamiliar model to unseen BookMIA data, while 7-epoch pre-training on them defines the familiar model.}, and observe clear differences in magnitude, location, and concentration, as illustrated in Figure~\ref{fig:motivation1} and Section~\ref{sec:motivation}. (1) \textbf{Decay of update magnitude.} Figure~\ref{fig:motivation1}(a,b) shows that parameter updates are strongly attenuated as data grows familiar, consistent with loss-convergence theory~\cite{bottou2010large,ruder2016overview}. (2)\textbf{Gradually stabilizing of update locations.} Figure~\ref{fig:motivation1}(c,d) shows that parameter updates evolve from broad activation regions to a stable core set of neurons, consistent with the sparse activation behavior of LLMs~\cite{liu2024probing}. (3)\textbf{Increasing update sparsity.} Figure~\ref{fig:motivation1}(a,b) and Figures~\ref{fig:motivation_four_features}(b,c) show that during training, an increasing fraction of updates concentrates in the top 10\% of neurons, consistent with Hessian spectrum reshaping during loss convergence~\cite{gur2018gradient}. 
Familiar data yields sparse, concentrated updates, whereas unfamiliar data leads to more distributed updates. Therefore, we can utilize these training dynamics and phase-shifted updates to support more generalizable pre-training data detection.

In this paper, we propose GDS, a novel pre-training data detection method that infers pre-training data by probing Gradient Deviation Scores of target samples without fine-tuning. GDS exploits gradient differences between member and non-member samples to train a lightweight classifier with strong generalization. Specifically, within the LoRA framework, we collect per-sample gradients across layers during backpropagation, encoding them as eight-dimensional features that capture the magnitude, location, and concentration of updates. Compared to non-members, member samples show smaller matrix-level and row-wise update magnitudes, lower FFN but higher attention eccentricity, greater variability in matrix-level and row-wise updates, a higher share of top gradients, and fewer sparse neurons. Then, we extract discriminative statistics from the signals to build fixed-dimensional features, which are fed to a lightweight MLP for binary membership classification.\footnote{https://anonymous.4open.science/r/emnlp-pdd-7C1A/} Our contributions are as follows:

\begin{itemize}
    \item From a training optimization perspective, we analyze how LLMs evolve from unfamiliarity to familiarity with data and propose using stage-wise parameter update dynamics to identify pre-training data, offering a novel direction to solve this task.
    \item We propose gradient deviation features capturing magnitude, location, and concentration, and introduce a gradient deviation score–based method for pre-training data detection without fine-tuning.
    \item Extensive experiments across diverse datasets and backbone models validate the effectiveness and generalization of GDS, while interpretability analyses and semi-supervised results offer a practical way to data detection.
\end{itemize}
\section{Related Work}

\paragraph{Unsupervised Likelihood-Based Methods.}
Early works distinguish member and non-member samples by capturing static statistical metrics in token-level likelihood distributions. Global-likelihood–based methods, such as PPL~\cite{li2025estimating}, Zlib~\cite{carlini2021extracting}, and Lowercase~\cite{carlini2021extracting}, perform discrimination by exploiting global probability features, compression entropy, and case-conversion ratios. However, global likelihood is highly sensitive to word-frequency effects, leading to unstable performance. Min-k\%~\cite{shi2024detecting} and Min-k\%++~\cite{zhang2025min} mitigate global dependence by focusing on low-probability outlier tokens or target–candidate probability comparisons, with k and text length jointly determining the analyzed token subset. However, the fixed choice of k lacks adaptivity, causing severe score fluctuations in short texts with low-frequency words. Although PC-PDD~\cite{zhang2024pretraining} approximates the word-frequency distribution of pretraining corpora using public datasets, these references lack fidelity due to limited domain coverage. These unsupervised methods are effective in practice, but they often hit a performance ceiling in complex scenarios. 

\paragraph{Supervised Fine-tuning Methods.}
Recent studies~\cite{zhangfine, choicontaminated} adopt lightweight fine-tuning to actively amplify asymmetric differences between member and non-member samples.
KDS~\cite{choicontaminated} evaluates dataset contamination by comparing kernel similarity matrices of embeddings before and after fine-tuning, assuming non-member datasets undergo larger embedding changes than member datasets. However, it only estimates dataset-level contamination and cannot reliably detect contamination at the sample-level. Another representative method FSD~\cite{zhangfine} detects membership by comparing loss before and after fine-tuning, assuming non-member samples experience a larger loss reduction than member samples. Nevertheless, these methods assume that fine-tuning data closely matches the target distribution, requiring additional tuning on similar non-members, limiting cross-dataset generalization.
\section{Motivation}
\label{sec:motivation}


Inspired by optimization theory~\cite{ruder2016overview, bottou2010large}, we investigate how LLMs shift from unfamiliarity to familiarity with data and how this process relates to the dynamics of model parameters during training. To this end, we fine-tune LLaMA-7B with LoRA on the unseen split of BookMIA and track LoRA updates over 7 epochs. Stage-wise analysis in Figure~\ref{fig:motivation_four_features} shows the evolution of gradient features. We observe that models familiar with test data exhibit smaller, sparser, and more stable gradient updates, which motivates our Gradient Deviation Score for membership detection.

\subsection{Decay of Update Magnitude}
Model training aims to minimize the loss function $L(\theta)$, with update magnitude determined by the gradient norm $\|\nabla L(\theta)\|$. According to optimization convergence theory~\cite{ruder2016overview}, as the parameters $\theta_t$ approach the optimum $\theta^*$, the gradient norm of the loss function vanishes, i.e., \( \lim_{\theta \to \theta^*} \|\nabla L(\theta_t)\| = 0 \). This causes the parameter update magnitude to decrease synchronously. Therefore, we use $\Delta\theta_t$ as the core indicator of the \textbf{average update magnitude} across all trainable parameters at iteration $t$:
\begin{equation}
\Delta\theta_t = \frac{1}{N} \sum_{i=1}^{N} \left| \theta_{t,i} - \theta_{t-1,i} \right|,
\label{eq:mean_update_magnitude_basic}
\end{equation}
where $N$ denotes the total number of parameters.


As training progresses, both the loss $L(\theta_t)$ and gradient norm $\|\nabla L(\theta_t)\|$ decrease, reducing the update magnitude $\Delta\theta_t$. Eventually, $L(\theta_t)$ stabilizes near its minimum, parameter updates $\Delta\theta_t$ become negligible. As shown in Figure~\ref{subfig:mean_trend}, the mean update magnitude decreases monotonically with training, with the largest changes occurring in the first two epochs, consistent with loss convergence.

\begin{figure*}[!t]
    \centering
    \includegraphics[width=0.85\textwidth]{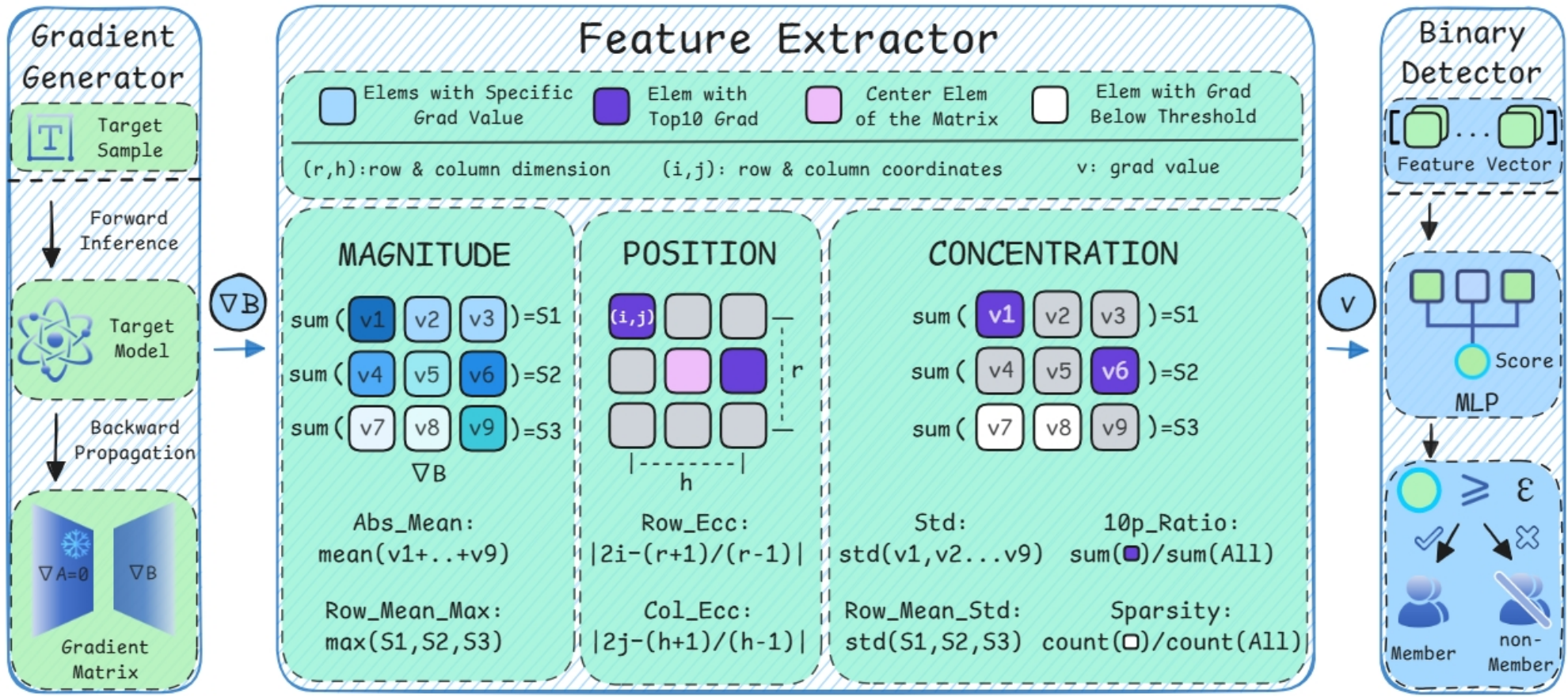}
    \caption{Overview of Gradient Deviation Scores method.}
    \label{fig:framework}
    \vspace{-12pt}
\end{figure*}

\subsection{Gradually Stabilizing of Update Locations}
Studies~\cite{liu2024probing, tang2025identifying} have shown that neural activations occur at different locations when processing member and non-member samples. Inspired by this, we adopt the \textbf{parameter update eccentricity} as the core indicator, which quantifies the deviation of the top-10\% parameter update positions from the global parameter centroid. We first define the global centroid as:
\begin{equation}
\theta_c = \frac{1}{N} \sum_{i=1}^N \theta_i,
\label{eq:param_center}
\end{equation}
where $\theta_i$ is the \textbf{coordinate} of the $i$-th parameter.

The update eccentricity at iteration $t$ is then defined as:
\begin{equation}
E_t = \frac{1}{\operatorname{card}(U_{10,t})} \sum_{\theta_i \in U_{10,t}} \left\| \theta_i - \theta_c \right\|_2,
\label{eq:update_eccentricity}
\end{equation}
where $\operatorname{card}(\cdot)$ denotes the number of elements in a set, $U_{10,t} = \{ \|\Delta\theta_{t,i}\|_1 \in \text{Top10\%}(\|\Delta\theta_t\|_1)\}$, $\|\cdot\|_1$ denotes the $\ell_1$ norm and $\|\cdot\|_2$ denotes the $\ell_2$ norm.

In the initial training stage, activation patterns are unstable, leading to scattered $U_{10,t}$ and thus random large values of $E_t$. As training proceeds, the model identifies core parameters related to data features: $\theta_t$ stabilizes to the optimal subset $\theta^*$, and $U_{10,t}$ converges to the core parameter region, resulting in $E_t$ gradually stabilizing at a low level. Figure~\ref{subfig:other_trend} shows that the eccentricity of parameter update evolves to a relatively stable value, consistent with the evolution of the neuron activation pattern, suggesting that training identifies the core parameters and stabilizes updates around them.

\subsection{Increasing Update Sparsity}

The Hessian spectrum~\cite{gur2018gradient, frankleearly} characterizes the curvature of the loss landscape, which governs the direction and efficiency of parameter updates. For loss $L(\theta)$, its Hessian matrix $\mathcal{H}(\theta) = \nabla^2 L(\theta)$ admits the eigenvalue decomposition:
\begin{equation}
\mathcal{H}(\theta) = U\Lambda U^T,
\label{eq:hessian_eig_decomp}
\end{equation}
where $U$ and $\Lambda$ denote the orthogonal eigenvectors and eigenvalues, respectively. During training, the Hessian spectrum evolves from nearly uniform eigenvalues to a structure dominated by a few large values with most approaching zero.

This reshaping yields concentrated parameter updates, focusing on directions with large eigenvalues while leaving most others nearly unchanged, motivating our two joint indicators. $S_t$ measures the fraction of parameters with negligible updates ($<10^{-6}$) at iteration $t$, and $T_{10,t}$ denotes the fraction of total update magnitude contributed by the top 10\% largest updates.
\begin{align}
    S_t &= \frac{\operatorname{card}\left({i \mid |\Delta\theta_{t,i}| < 10^{-6}}\right)}{N},\\
    T_{10,t} &= \frac{\sum_{|\Delta\theta_{t,i}| \in U_{10,t}} |\Delta\theta_{t,i}|}{||\Delta\theta_t||_1}.
\end{align}


As training progresses, parameter updates increasingly concentrate along core directions. Consequently, $S_t$ gradually increases, while $T_{10,t}$ remains relatively large and stable(0.26–0.27), indicating that update energy is consistently dominated by a small subset of parameters. Figures~\ref{subfig:sparsity_trend} and~\ref{subfig:eccentricity_trend} confirm this growing sparsity and stable Top 10\% contribution. Further details on the variation trends of the three metrics are provided in the appendix~\ref{app:motivation_experiment}.
\section{Method}

\subsection{Task Definition}
Given a target text $x$ and a target LLM $f_\theta$ pre-trained on a corpus $D$, the goal of pre-training data detection task is to construct a detector $h(x, f_\theta)$ to determine whether $x$ belongs to $D$. This task can be formalized as a binary classification problem: 
\(h(x, f_\theta) \rightarrow \{0, 1\}\)
where ``1'' indicates $x \in D$ as member and ``0'' indicates $x \notin D$ as non-member. 

As illustrated in Figure~\ref{fig:framework}, GDS comprises three stages: gradient matrix acquisition, feature vector extraction and light Multilayer Perceptron(MLP) training. For each training sample, we perform inference and backpropagation to obtain LoRA gradients across $l$ layers and $m$ sub-modules, yielding $l \times m$ gradient matrices $\mathbb{G}$. From these matrices, we extract eight-dimensional features to form a gradient feature vector $\mathbb{V}^{l \times m \times 8}$, which is used to train a MLP classifier for binary membership prediction. During inference, the same feature extraction process is applied to a target sample $x^*$, and the resulting feature vector is fed into the trained MLP to produce the final classification result.

\subsection{Gradient Matrix Acquisition}
Given the target model $f_\theta$ and training sample $x$, we first initialize $f_\theta$ with LoRA, yielding $l \times m$ LoRA matrices $\mathbb{L}$. We then feed a single sample $x$ into $f_\theta$ for forward inference and backpropagation, obtaining LoRA gradient matrices $\mathbb{G} = \nabla_{\mathbb{L}} \mathcal{L}_{\text{CLM}}(f_\theta(x))$, where $\mathcal{L}_{\text{CLM}}$ denotes the causal language modeling loss of the pre-trained LLM. Since the LoRA\_B matrices are fully initialized to zero, a single round of gradient propagation leads to zero gradients for the LoRA\_A matrices. Thus, we only collect the gradient matrices of LoRA\_B for subsequent processing.

\subsection{Feature Vector Extraction}
Given the gradient matrices $\mathbb{G}$, we extract features based on the three parameter update trends introduced in Section~\ref{sec:motivation}.

\subsubsection{Magnitude}
We propose two magnitude indicators to measure the overall scale of gradient updates in the LoRA parameter space.

\textbf{(1) Absolute Mean}: The mean of the absolute values of all elements in the gradient matrix, which characterizes the overall parameter update strength:
\begin{equation}
\text{Abs\_Mean} = \frac{1}{r \cdot h} \sum_{i=1}^r \sum_{j=1}^h |\mathbb{G}_{i,j}|,
\label{eq:mean_gradient_magnitude}
\end{equation}
where \( \mathbb{G}_{i,j} \) denotes the element in the \( i \)-th row and \( j \)-th column of the gradient matrix with $r$ rows and $h$ columns.

\textbf{(2) Max Row Mean}: The mean of each row in the LoRA matrix and take the maximum value, which characterizes the most sample-responsive local optimal response dimension with the largest parameter update magnitude:
\begin{equation}
\text{Row\_Mean\_Max} = \max_{1 \leq i \leq r} \left( \frac{1}{h} \sum_{j=1}^h |\mathbb{G}_{i,j}| \right).
\label{eq:max_row_gradient_mean}
\end{equation}

\subsubsection{Position}
Screen the top 10\% of gradient elements by absolute value from $\mathbb{G}$ and calculate their offset from this matrix center, capturing offset characteristics of core gradient positions for \textbf{(1) Row Eccentricity} and \textbf{(2) Column Eccentricity}, respectively:
\begin{equation}
\text{Row\_Ecc} = \frac{1}{\operatorname{card}(S)} \sum_{(i,j) \in S} \left| \frac{2i - (r+1)}{r-1} \right|,
\label{eq:top10_grad_i_bias}
\end{equation}
\begin{equation}
\text{Col\_Ecc} = \frac{1}{\operatorname{card}(S)} \sum_{(i,j) \in S} \left| \frac{2j - (h+1)}{h-1} \right|,
\label{eq:top10_grad_j_bias}
\end{equation}
where \( S = \{(i,j) \mid |\mathbb{G}_{i,j}| \in \text{Top10\%}(|\mathbb{G}|)\} \) denotes the set of indices for the top 10\% of elements in the gradient matrix \( \mathbb{G} \) ranked by absolute gradient value. Top gradients near the center correspond to an eccentricity of 0, while those near the edge correspond to a score of 1.

\subsubsection{Concentration}
We propose four indicators to measure the concentration of gradient distributions.

\textbf{(1) Top-10\% Ratio}: The ratio of the sum of the top 10\% largest gradient magnitudes to the total update magnitude of the gradient matrix, which quantifies the contribution ratio of core gradients:
\begin{equation}
\text{10p\_Ratio} = \frac{\sum_{(i,j) \in S} |\mathbb{G}_{i,j}|}{\|\mathbb{G}\|_1},
\label{eq:grad_top10p_ratio}
\end{equation}

\textbf{(2) Sparsity}: The proportion of gradient elements with absolute values less than $10^{-6}$, quantifying the sparsity degree of gradient distributions:
\begin{equation}
\text{Sparsity} = \frac{\operatorname{card}\left(\{(i,j) \mid |\mathbb{G}_{i,j}| < 10^{-6}\}\right)}{r \cdot h}.
\label{eq:grad_sparsity}
\end{equation}

\textbf{(3) Standard Deviation}: The standard deviation of the elements in the gradient matrix, which quantifies the dispersion degree of parameter updates:
\begin{equation}
\text{Std} = \sqrt{\frac{1}{r \cdot h} \sum_{i=1}^r \sum_{j=1}^h \left(|\mathbb{G}_{i,j}| - \text{Abs\_mean}\right)^2}.
\label{eq:gradient_std}
\end{equation}

\begin{table*}[!t]
\centering
\scriptsize
\caption{Comparison(AUROC/TPR@5\%FPR) on 4 Datasets across 5 target models. Target models are denoted as 2.7B/6B/6.7B/6.9B/7B, corresponding to Neo-2.7B/GPT-J-6b/OPT-6.7b/Pythia-6.9b/LLaMA-7B respectively.}
\label{tab:main_performance_two_rows}
\setlength{\tabcolsep}{3.2pt}
\renewcommand{\arraystretch}{1}

\begin{tabular}{c @{\hspace{2pt}} ccccc @{\hspace{7pt}} ccccc}
\toprule
\multicolumn{1}{c}{} & \multicolumn{5}{c}{WikiMIA} & \multicolumn{5}{c}{ArXivTection} \\
\cmidrule(l{2pt} r{2pt}){2-6} \cmidrule(l{2pt} r{2pt}){7-11}
\multirow{-2}{*}{\makecell{Method}} & 2.7B & 6B & 6.7B & 6.9B & 7B & 2.7B & 6B & 6.7B & 6.9B & 7B \\
\midrule
PPL & 0.61/0.13 & 0.64/0.13 & 0.60/0.12 & 0.63/0.13 & 0.69/0.14 & 0.62/0.13 & 0.64/0.13 & 0.60/0.12 & 0.63/0.13 & 0.70/0.14 \\
ZLib & 0.58/0.10 & 0.60/0.11 & 0.58/0.10 & 0.59/0.10 & 0.71/0.22 & 0.58/0.10 & 0.60/0.11 & 0.58/0.09 & 0.59/0.11 & 0.71/0.22 \\
Min-k & 0.65/0.17 & 0.67/0.19 & 0.63/0.15 & 0.67/0.19 & 0.73/0.18 & 0.65/0.17 & 0.68/0.19 & 0.63/0.15 & 0.67/0.19 & 0.72/0.18 \\
Min-k++ & 0.67/0.15 & 0.69/0.19 & 0.65/0.11 & 0.70/0.18 & 0.82/0.22 & 0.67/0.15 & 0.69/0.19 & 0.65/0.10 & 0.70/0.17 & 0.83/0.22 \\
FSD & \textbf{0.92/0.77} & \textbf{0.95/0.78} & 0.90/0.63 & 0.90/\textbf{0.66} & 0.92/0.41 & 0.91/0.65 & 0.96/0.79 & 0.89/0.63 & 0.95/0.66 & 0.94/0.81 \\
Ours & 0.90/0.60 & 0.93/0.66 & \textbf{0.94/0.67} & \textbf{0.92}/0.63 & \textbf{0.96/0.84} & \textbf{0.94/0.73} & \textbf{0.97/0.86} & \textbf{0.94/0.75} & \textbf{0.95/0.83} & \textbf{0.97/0.85} \\
\bottomrule

\multicolumn{1}{c}{} & \multicolumn{5}{c}{BookTection} & \multicolumn{5}{c}{BookMIA} \\
\cmidrule(l{2pt} r{2pt}){2-6} \cmidrule(l{2pt} r{2pt}){7-11}
\multirow{-2}{*}{\makecell{Method}} & 2.7B & 6B & 6.7B & 6.9B & 7B & 2.7B & 6B & 6.7B & 6.9B & 7B \\
\midrule
PPL & 0.69/0.15 & 0.74/0.25 & 0.64/0.13 & 0.73/0.25 & 0.71/0.25 & 0.29/0.02 & 0.22/0.13 & 0.23/0.01 & 0.59/0.19 & 0.56/0.21 \\
ZLib & 0.56/0.16 & 0.58/0.20 & 0.55/0.14 & 0.58/0.19 & 0.57/0.19 & 0.20/0.02 & 0.38/0.13 & 0.15/0.00 & 0.49/0.17 & 0.48/0.18 \\
Min-k & 0.71/0.18 & 0.75/0.27 & 0.67/0.15 & 0.74/0.26 & 0.71/0.25 & 0.45/0.05 & 0.60/0.20 & 0.40/0.03 & 0.62/0.20 & 0.60/0.21 \\
Min-k++ & 0.64/0.13 & 0.67/0.18 & 0.61/0.11 & 0.66/0.19 & 0.63/0.15 & 0.54/0.13 & 0.64/0.29 & 0.49/0.09 & 0.60/0.23 & 0.58/0.20 \\
FSD & 0.92/0.53 & 0.91/0.52 & 0.96/0.77 & 0.93/0.59 & 0.92/0.55 & 0.98/0.93 & 0.97/0.89 & 0.98/0.96 & 0.98/0.93 & 0.98/0.91 \\
Ours & \textbf{0.96/0.84} & \textbf{0.97/0.88} & \textbf{0.98/0.92} & \textbf{0.96/0.83} & \textbf{0.98/0.92} & \textbf{0.99/0.98} & \textbf{0.99/0.99} & \textbf{0.99/0.99} & \textbf{0.99/0.98} & \textbf{0.99/0.99} \\
\bottomrule
\end{tabular}
\end{table*}

\textbf{(4) Row Mean Standard Deviation}: The mean of each row in the LoRA matrix and take the standard deviation, which quantifies the consistency of update strength across all rows:
\begin{equation}
\text{Row\_Mean\_Std} = \sqrt{\frac{1}{r} \sum_{i=1}^r \left( \bar{\mathbb{G}}_i - \mu_{\bar{\mathbb{G}}} \right)^2},
\label{eq:row_gradient_mean_std}
\end{equation}
where $\bar{\mathbb{G}}_i = \frac{1}{h} \sum_{j=1}^h |\mathbb{G}_{i,j}|$ denotes the $i$-th row's absolute mean, $\mu_{\bar{\mathbb{G}}} = \frac{1}{r} \sum_{i=1}^r \bar{\mathbb{G}}_i$ denotes the mean of row means.


We compute the aforementioned eight feature values from all gradient matrices to derive a feature vector $\mathbb{V}^{l \times m \times 8}$ per sample, which is then used for subsequent MLP training and inference.

\subsection{Light MLP Training}
Given the feature vectors $\mathbb{V}_i$ from all training samples vectors $\mathbb{V}_s$, we feed them into a lightweight MLP for training. The target output corresponds to the binary label $\{0, 1\}$ for each sample, where $0$ denotes non-member and $1$ denotes member. The training loss is defined as follows:
\begin{equation}
\mathcal{L} = -\frac{1}{N}\sum_{i=1}^{N}\left[y_i\log\hat{y}_i + (1-y_i)\log(1-\hat{y}_i)\right],
\end{equation}
where $N$ denotes the number of training samples, $y_i\in\{0,1\}$ is the ground-truth label of the $i$-th sample, and $\hat{y}_i$ is the predicted probability output by the MLP.

\begin{table*}[!t]
\centering
\scriptsize
\caption{Performance Comparison on Mimir datasets with 7 subsets across 5 target models using AUROC scores.}
\label{tab:mimir_main_results_no_wrap_6methods}
\setlength{\tabcolsep}{2pt}
\renewcommand{\arraystretch}{1}

\begin{tabular}{c @{\hspace{2pt}} ccccc @{\hspace{2pt}} ccccc @{\hspace{2pt}} ccccc @{\hspace{2pt}} ccccc}
\toprule
\multicolumn{1}{c}{} & \multicolumn{5}{c}{Wikipedia} & \multicolumn{5}{c}{Github} & \multicolumn{5}{c}{Pile CC} & \multicolumn{5}{c}{PubMed Central} \\
\cmidrule(l{3pt} r{3pt}){2-6} \cmidrule(l{3pt} r{3pt}){7-11} \cmidrule(l{3pt} r{3pt}){12-16} \cmidrule(l{3pt} r{3pt}){17-21}
\multirow{-2}{*}{\makecell{Method}} & 2.7B & 6B & 6.7B & 6.9B & 7B & 2.7B & 6B & 6.7B & 6.9B & 7B & 2.7B & 6B & 6.7B & 6.9B & 7B & 2.7B & 6B & 6.7B & 6.9B & 7B \\
\midrule
PPL & 0.56 & 0.57 & 0.56 & 0.56 & 0.56 & 0.78 & 0.79 & 0.78 & 0.90 & 0.83 & 0.56 & 0.57 & 0.56 & 0.56 & 0.52 & 0.78 & 0.79 & 0.78 & 0.78 & 0.72 \\
ZLib & 0.62 & 0.65 & 0.59 & 0.65 & 0.58 & \textbf{0.90} & 0.91 & 0.79 & 0.91 & 0.86 & 0.55 & 0.55 & 0.55 & 0.55 & 0.52 & 0.78 & 0.78 & 0.73 & 0.77 & 0.71 \\
Min-k & \textbf{0.66} & \textbf{0.68} & \textbf{0.65} & 0.68 & 0.63 & 0.88 & 0.89 & 0.76 & 0.90 & 0.83 & 0.56 & 0.57 & 0.56 & 0.56 & 0.52 & 0.81 & 0.80 & 0.74 & \textbf{0.79} & 0.72 \\
Min-k++ & 0.65 & \textbf{0.68} & 0.60 & \textbf{0.69} & 0.56 & 0.84 & 0.86 & 0.56 & 0.86 & 0.73 & 0.54 & 0.55 & 0.54 & 0.56 & 0.51 & 0.71 & 0.72 & 0.60 & 0.70 & 0.59 \\
FSD & 0.61 & 0.67 & 0.60 & 0.65 & 0.60 & 0.77 & 0.80 & 0.62 & 0.77 & 0.72 & 0.55 & 0.55 & 0.54 & 0.55 & 0.52 & 0.71 & 0.79 & 0.63 & 0.56 & 0.63 \\
Ours & 0.63 & 0.64 & \textbf{0.65} & 0.65 & \textbf{0.64} & 0.90 & \textbf{0.92} & \textbf{0.88} & \textbf{0.92} & \textbf{0.91} & \textbf{0.59} & \textbf{0.59} & \textbf{0.59} & \textbf{0.57} & \textbf{0.59} & \textbf{0.84} & \textbf{0.85} & \textbf{0.84} & \textbf{0.79} & \textbf{0.82} \\
\bottomrule

\multicolumn{1}{c}{} & \multicolumn{5}{c}{ArXiv} & \multicolumn{5}{c}{DM Mathematics} & \multicolumn{5}{c}{HackerNews} & \multicolumn{5}{c}{Average} \\
\cmidrule(l{3pt} r{3pt}){2-6} \cmidrule(l{3pt} r{3pt}){7-11} \cmidrule(l{3pt} r{3pt}){12-16} \cmidrule(l{3pt} r{3pt}){17-21}
\multirow{-2}{*}{\makecell{Method}} & 2.7B & 6B & 6.7B & 6.9B & 7B & 2.7B & 6B & 6.7B & 6.9B & 7B & 2.7B & 6B & 6.7B & 6.9B & 7B & 2.7B & 6B & 6.7B & 6.9B & 7B \\
\midrule
PPL & \textbf{0.78} & \textbf{0.79} & 0.65 & \textbf{0.78} & 0.70 & 0.78 & 0.79 & 0.65 & 0.91 & 0.31 & 0.65 & \textbf{0.62} & 0.59 & \textbf{0.62} & \textbf{0.59} & 0.64 & 0.66 & 0.62 & 0.73 & 0.59 \\
ZLib & 0.77 & 0.78 & 0.68 & 0.77 & 0.71 & 0.82 & 0.81 & 0.80 & 0.81 & 0.23 & \textbf{0.68} & 0.60 & 0.58 & 0.60 & 0.58 & 0.65 & 0.67 & 0.63 & 0.72 & 0.58 \\
Min-k & \textbf{0.78} & \textbf{0.79} & 0.65 & \textbf{0.78} & 0.70 & 0.93 & 0.93 & 0.92 & 0.92 & 0.32 & 0.65 & \textbf{0.62} & 0.59 & \textbf{0.62} & \textbf{0.59} & 0.71 & 0.73 & 0.69 & 0.75 & 0.60 \\
Min-k++ & 0.60 & 0.64 & 0.53 & 0.70 & 0.57 & 0.77 & 0.79 & 0.67 & 0.75 & 0.22 & 0.53 & 0.57 & 0.51 & 0.60 & 0.53 & 0.60 & 0.63 & 0.56 & 0.67 & 0.53 \\
FSD & 0.72 & 0.78 & 0.55 & 0.70 & 0.52 & 0.58 & 0.85 & 0.60 & 0.74 & 0.53 & 0.61 & 0.60 & 0.57 & 0.56 & 0.51 & 0.61 & 0.66 & 0.59 & 0.62 & 0.55 \\
Ours & \textbf{0.78} & \textbf{0.79} & \textbf{0.78} & \textbf{0.78} & \textbf{0.77} & \textbf{0.95} & \textbf{0.95} & \textbf{0.94} & \textbf{0.95} & \textbf{0.95} & 0.58 & 0.60 & \textbf{0.60} & 0.57 & 0.58 & \textbf{0.73} & \textbf{0.75} & \textbf{0.74} & \textbf{0.76} & \textbf{0.70} \\
\bottomrule
\end{tabular}
\end{table*}

\section{Experiments}

\subsection{Experimental Setup}

\textbf{Datasets}
    \label{subsec:datasets}
Following prior work, we evaluate on five prevalent datasets: WikiMIA, BookMIA~\cite{shi2024detecting},ArxivTection, BookTection~\cite{duartecop}, and MIMIR~\cite{duanmembership}, a widely recognized challenging benchmark for pretraining data detection.

\textbf{Target Models}
    \label{subsec:target_models}
We evaluate five open-source LLMs with diverse architectures: Neo-2.7B~\cite{black2021gpt}, GPT-J-6B~\cite{wang2021gpt}, OPT-6.7B~\cite{zhang2022opt}, Pythia-6.9B~\cite{biderman2023pythia}, and LLaMA-7B~\cite{touvron2023llama} from Hugging Face.

\textbf{Comparison Methods}
    \label{subsec:baselines}
We select five representative state-of-the-art baselines. Four scoring function–based methods include PPL~\cite{carlini2021extracting}, ZLib~\cite{carlini2021extracting}, Min-k~\cite{shi2024detecting}, and Min-k++~\cite{zhang2025min}. The fine-tuning–enhanced method includes FSD~\cite{zhangfine}. More details are shown in Appendix~\ref{app:baseline_details}.

\textbf{Evaluation Metrics}
    \label{subsec:metrics}
Two metrics evaluate the binary pre-training data detection task: \textbf{AUROC} measures overall discriminative performance, and \textbf{TPR@5\%FPR} captures practical effectiveness by assessing detection rate under a 5\% false positive constraint.

\textbf{Implementation Details}
We use the PEFT library~\cite{mangrulkar2022peft} for LoRA-based~\cite{hu2022lora} gradient feature extraction without parameter updates. The MLP is trained on 30\% of the data, with the remaining 70\% used for inference evaluation under FSD settings. More Details can be seen in Appendix~\ref{app:experiment_setting}.

\begin{table}[!t]
\centering
\footnotesize
\caption{Ablation results using AUROC on WikiMIA and ArxivTection. - denotes the features removal. }
\label{tab:feature_category_ablation}
\setlength{\tabcolsep}{6pt}
\renewcommand{\arraystretch}{1.1}
\begin{tabular}{l ccc}
\toprule
Dataset & -Magnitude & -Concentrate & -Position\\
\midrule
WikiMIA & 0.94 (↓0.02) & 0.94 (↓0.02) & 0.93 (↓0.03) \\
Arxiv. & 0.96 (↓0.01) & 0.95 (↓0.02) & 0.95 (↓0.02) \\
\bottomrule
\end{tabular}
\end{table}

\begin{table}[!t]
\centering
\footnotesize
\caption{Ablation results using AUROC and TPR@5\%FPR. ATT-Only refers to performance validation using only attention module.}
\label{tab:model_module_ablation}
\setlength{\tabcolsep}{9pt}
\renewcommand{\arraystretch}{1.1}
\begin{tabular}{lccc}
\toprule
Dataset & ATT-Only & FFN-Only & Origin \\
\midrule
WikiMIA & 0.94 / 0.68 & 0.93 / 0.64 & 0.96 / 0.84 \\
Arxiv. & 0.95 / 0.76 & 0.93 / 0.72 & 0.97 / 0.85 \\
\bottomrule
\end{tabular}
\end{table}

\subsection{Main Results}
\label{subsec:main_experiments}
Table~\ref{tab:main_performance_two_rows} reports results on WikiMIA, ArxivTection, BookTection, and BookMIA. On WikiMIA with LLaMA-7B, GDS reaches an AUC of 0.96, outperforming the best baseline, FSD, by 0.04 and substantially surpassing score-based methods such as Min-k++. GDS performs best in most settings, with particularly large TPR@5\%FPR gains on BookTection and BookMIA; on BookTection with LLaMA-7B, the improvement is nearly 67.3

Table~\ref{tab:mimir_main_results_no_wrap_6methods} reports results on seven MIMIR subsets. Despite performance drops caused by similar data distributions, GDS delivers the best average gains ({+2.8\%, +2.7\%, +7.2\%, +1.3\%, +16.6\%}) across models, with notable advantages on PubMed Central, DM Mathematics, and GitHub. With LLaMA, these three subsets reach AUCs of 0.82, 0.95, and 0.91. GDS also remains the most stable across models, with AUC variation mostly within 0.04, indicating strong cross-model generalization.

\subsection{Ablation Study}

We perform two ablation studies: (1) removing each feature category individually, and (2) using features from only the Attention or FFN module. Table~\ref{tab:feature_category_ablation} shows that all categories help, with Position Offset contributing most; the full feature set performs best, confirming their complementarity (see Appendix~\ref{app:ablation_sub}). Table~\ref{tab:model_module_ablation} shows that Attention features are more informative than FFN features, but both single-module variants lag behind the full model, indicating that Attention and FFN capture complementary gradient patterns and are best used together.

\subsection{Analysis}
\subsubsection{Semi-Supervised Setting}
To test GDS in label-free real-world settings, we combine it with the SOTA unsupervised method MinK++. MinK++ first filters WikiMIA samples using thresholds of $-1.4$ and $-4.6$, producing 310 pseudo-labeled samples for training GDS and FSD. Table~\ref{tab:semi_supervised} results show semi-supervised GDS outperforms unsupervised MinK++ and semi-supervised FSD, demonstrating excellent noise resistance and practical value. Notably, GDS has a distinct advantage in TPR@5\%FPR, reflecting strong positive sample recall under extremely low false positive tolerance. This indicates GDS can more accurately and robustly discriminate pre-trained data, suppress non-member high-score interference, select purer member samples in high-confidence regions, and thus has stronger deployment value in low-false-alarm, high-reliability scenarios.

\begin{table}[!t]
\centering
\small
\caption{Semi-supervised performance (AUROC / TPR@5\%FPR) on WikiMIA.}
\label{tab:semi_supervised}
\setlength{\tabcolsep}{7pt}
\renewcommand{\arraystretch}{1.1}
\begin{tabular}{lccc}
\toprule
              & MinK++    & FSD       & GDS       \\
\midrule
LLaMA-7B      & 0.82/0.22 & 0.79/0.134& 0.86/0.40 \\
Pythia-6.9B   & 0.70/0.18 & 0.76/0.18 & 0.80/0.34 \\
GPT-J-6B      & 0.69/0.19 & 0.75/0.14 & 0.79/0.26 \\
OPT-6.7B      & 0.65/0.11 & 0.70/0.11 & 0.81/0.37 \\
GPT-Neo-2.7B  & 0.67/0.15 & 0.76/0.17 & 0.78/0.26 \\
\bottomrule
\end{tabular}
\end{table}

\begin{figure*}[!t]
    \centering
    \setlength{\tabcolsep}{2pt}
    \begin{subfigure}{0.2\textwidth}
        \centering
        \includegraphics[width=\textwidth]{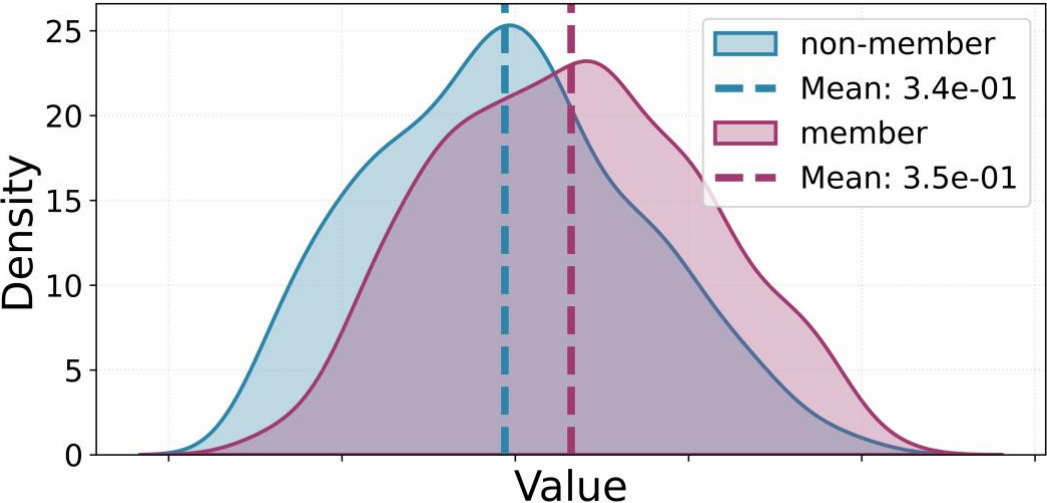}
        \caption{10p\_Ratio}
        \label{subfig:top10p_ratio}
    \end{subfigure}
    \hfill
    \begin{subfigure}{0.2\textwidth}
        \centering
        \includegraphics[width=\textwidth]{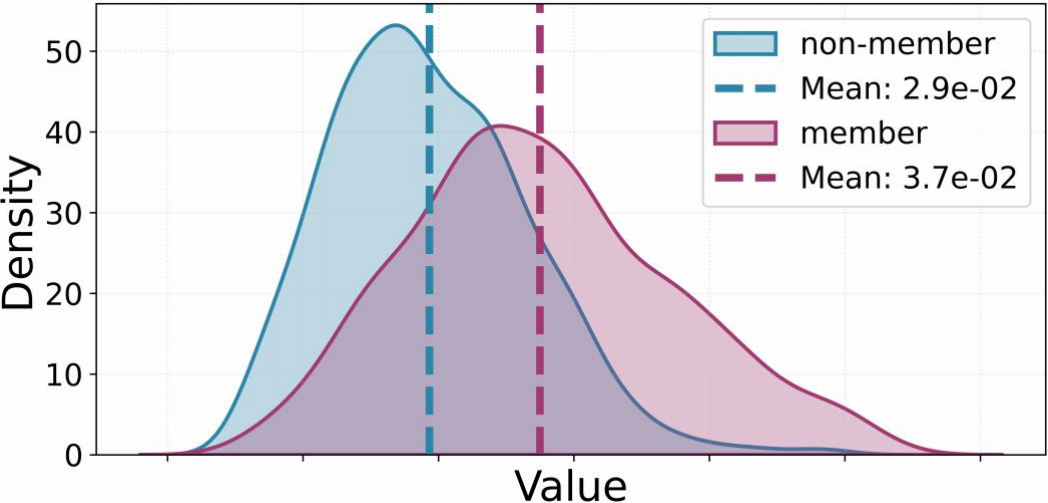}
        \caption{Sparsity}
        \label{subfig:sparsity}
    \end{subfigure}
    \hfill
    \begin{subfigure}{0.2\textwidth}
        \centering
        \includegraphics[width=\textwidth]{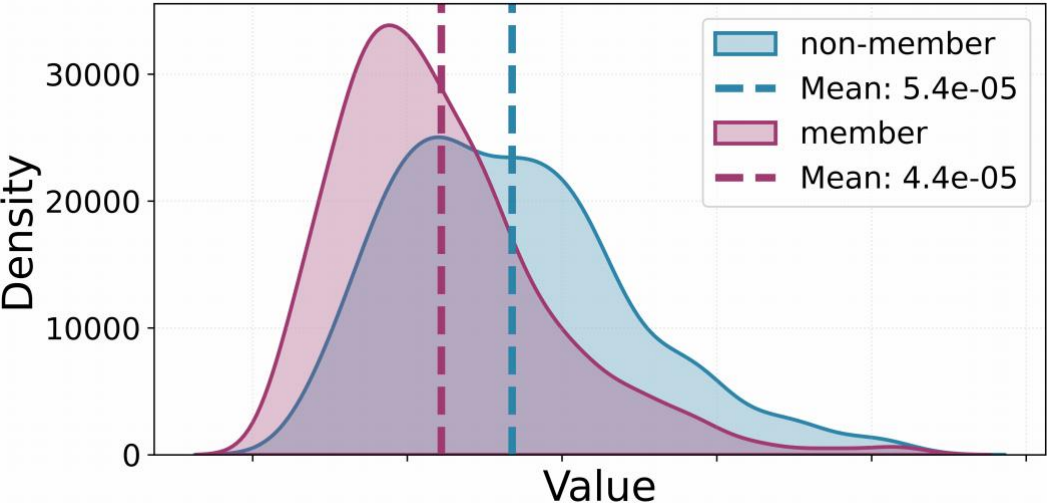}
        \caption{Abs\_Mean}
        \label{subfig:abs_mean}
    \end{subfigure}
    \hfill
    \begin{subfigure}{0.2\textwidth}
        \centering
        \includegraphics[width=\textwidth]{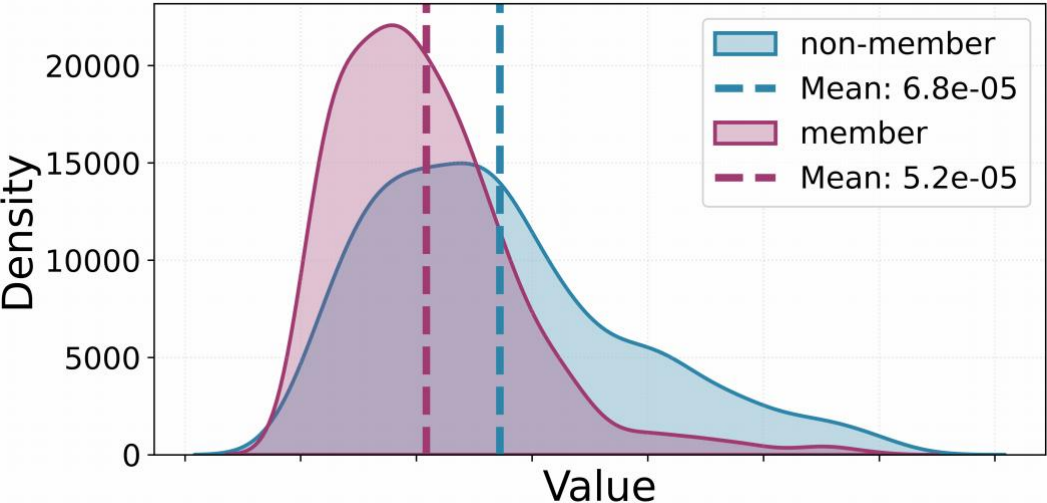}
        \caption{std}
        \label{subfig:std}
    \end{subfigure}
    
    \begin{subfigure}{0.2\textwidth}
        \centering
        \includegraphics[width=\textwidth]{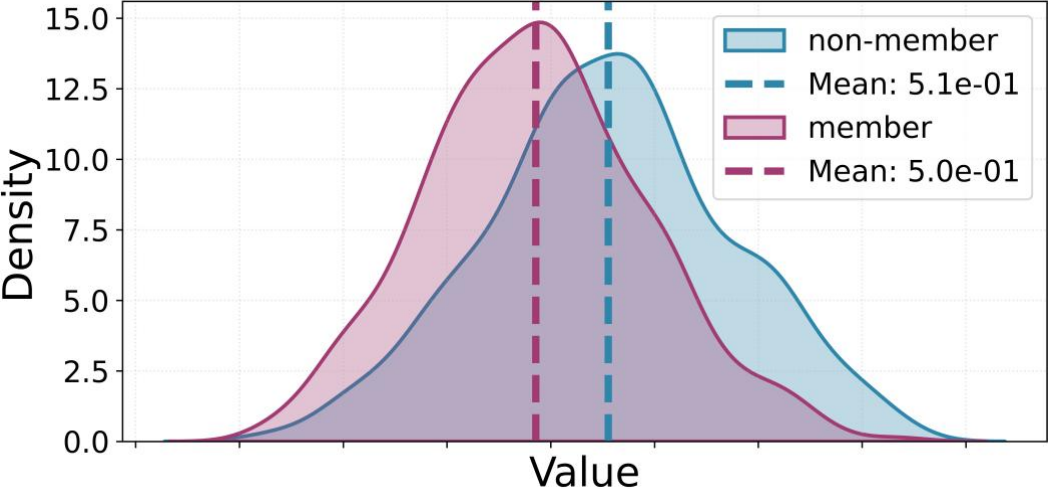}
        \caption{Row\_Ecc}
        \label{subfig:row_eccentricity}
    \end{subfigure}
    \hfill
    \begin{subfigure}{0.2\textwidth}
        \centering
        \includegraphics[width=\textwidth]{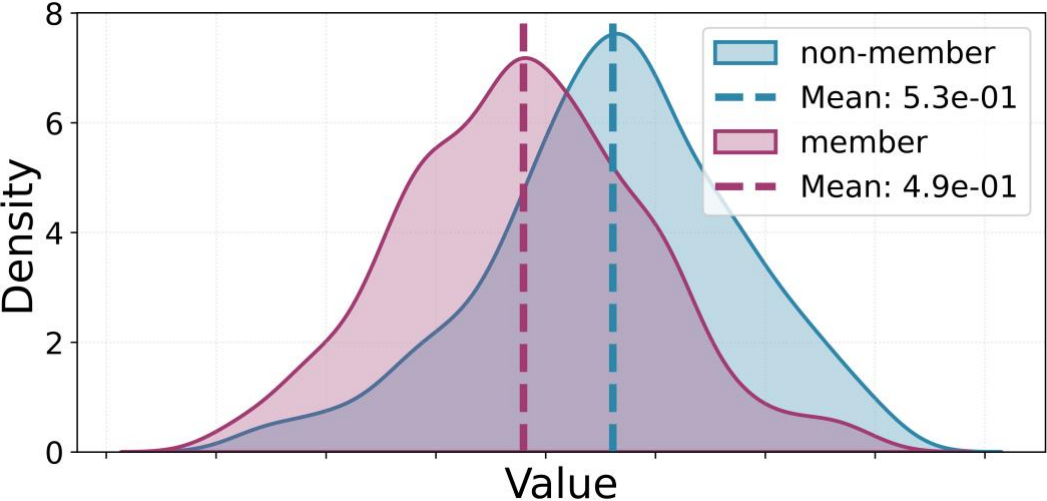}
        \caption{Col\_Ecc}
        \label{subfig:col_eccentricity}
    \end{subfigure}
    \hfill
    \begin{subfigure}{0.2\textwidth}
        \centering
        \includegraphics[width=\textwidth]{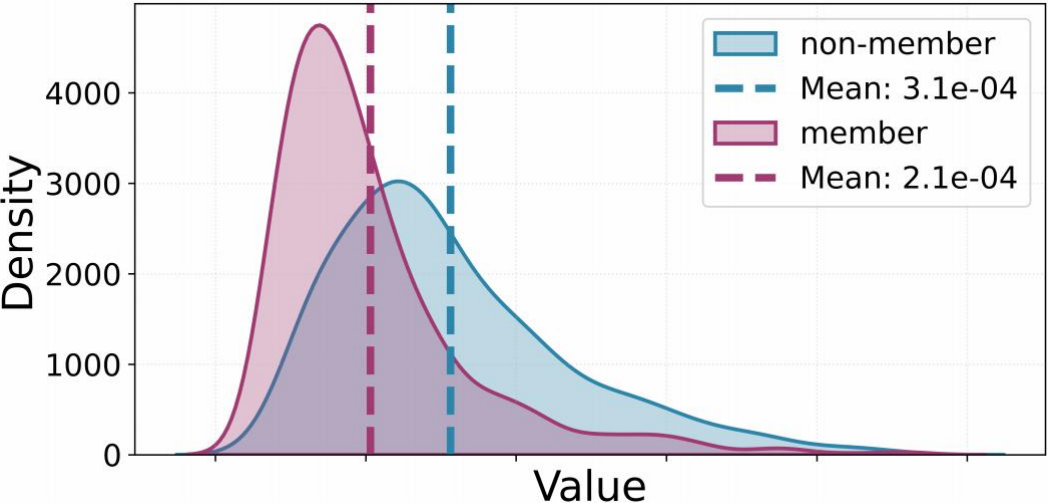}
        \caption{Row\_Mean\_Max}
        \label{subfig:row_mean_max}
    \end{subfigure}
    \hfill
    \begin{subfigure}{0.2\textwidth}
        \centering
        \includegraphics[width=\textwidth]{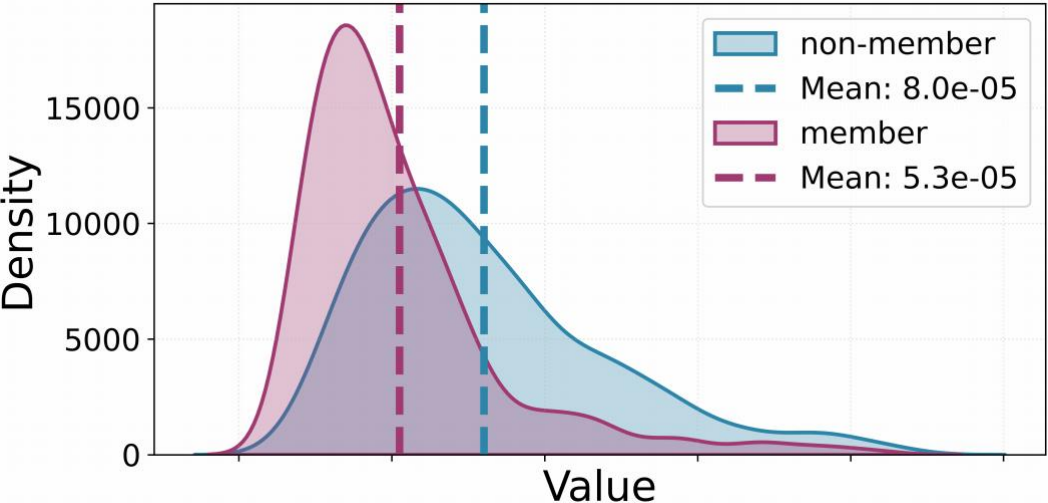}
        \caption{Row\_Mean\_Std}
        \label{subfig:std_row}
    \end{subfigure}
    
    \caption{Distribution differences of eight gradient features between member(red) and non-member(blue) samples. Dashed lines indicate distribution means. The x-axis shows feature values, and the y-axis shows probability density.}
    \label{fig:feature_distribution_k}
\end{figure*}

\subsubsection{Feature Distribution}
Figure~\ref{fig:feature_distribution_k} illustrates the eight features with the largest member–non-member discrepancies across all layers and sub-modules. Member samples show lower Abs\_Mean and Row\_Mean\_Max, higher Sparsity, lower Std and Row\_Mean\_Std, and higher 10p\_Ratio, indicating smaller, more concentrated, and more stable gradients closer to the propagation core. They also have lower Row\_Ecc and Col\_Ecc, suggesting that member-related parameters are more centrally located in weight matrices.

\begin{figure}[!t]
    \centering
    \includegraphics[width=0.40\textwidth]{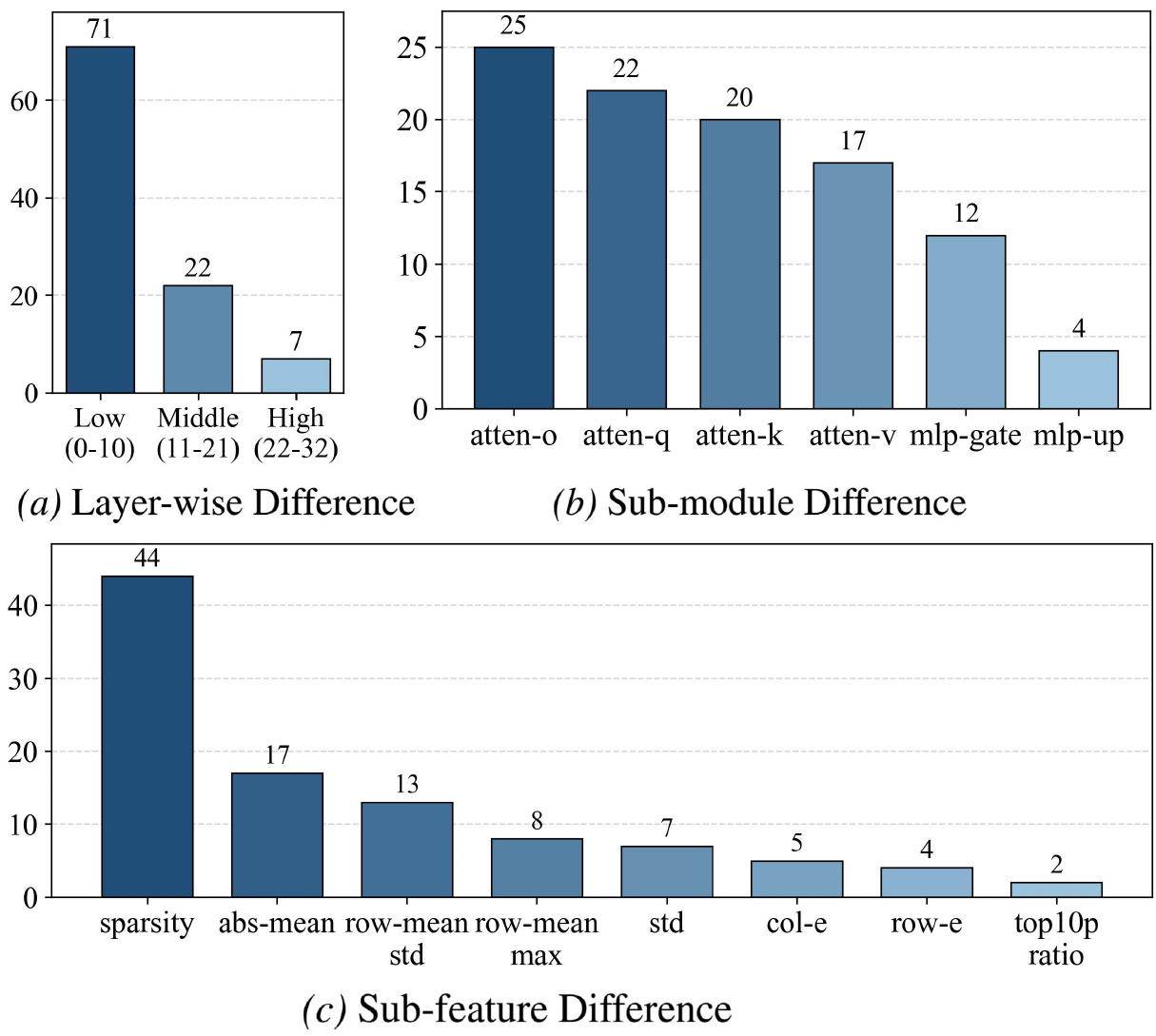}
    \caption{Distribution Difference Statistics of Discriminative Features. The y-axis shows the statistical count.}
    \label{fig:feature_difference_stats}
\end{figure}

Figure~\ref{fig:feature_difference_stats} shows that sub-features with the largest distribution differences are concentrated in lower layers, while middle and higher layers contribute far less. Attention-related sub-modules dominate the discriminative features, with clear variation across modules. Among gradient features, grad\_sparsity is by far the most discriminative.

\subsubsection{Transferability}
Although our method uses a unified feature extraction pipeline, it currently relies on dataset-specific classifiers.
We evaluate cross-dataset transfer between WikiMIA and ArxivTection, as well as a unified classifier trained on mixed data.
As shown in Table~\ref{tab:cross_dataset_generalization}, direct transfer suffers severe performance degradation due to dataset shift, while the unified classifier achieves strong results on combined data.
This indicates that our approach captures generalizable patterns related to pre-training familiarity.

\begin{table}[!t]
\centering
\small
\caption{Cross-Dataset Generalization Performance using AUROC. Wiki(arXiv) indicates training on ArxivTection with test on WikiMIA, while Mix indicates mixed datasets.}
\label{tab:cross_dataset_generalization}
\begin{tabular}{cccc}
\toprule
Method & Wiki (arXiv) & arXiv (Wiki) & Mix \\
\midrule
FSD    & 0.52         & 0.58         & 0.92 \\
Ours   & \textbf{0.66}& \textbf{0.68}& \textbf{0.95} \\
\bottomrule
\end{tabular}
\end{table}

\subsubsection{Efficiency Analysis}
We compare inference time and peak memory of PPL, FSD, and GDS on the first WikiMIA sample with LLaMA-7B in bfloat16. PPL takes 70 ms, FSD 760 ms, and GDS 1001 ms. Peak memory is about 13 GB for PPL and GDS, versus 16 GB for FSD. Although GDS is about 30\% slower than FSD due to gradient backpropagation, it performs no parameter updates, so it avoids optimizer states and matches the memory use of unsupervised methods while cutting memory by nearly 40\% to FSD. Furthermore, GDS maintains strong performance in low-resource scenarios and with reduced LoRA ranks(see Appendix~\ref{app:low_data} and ~\ref{app:sensitivity}), enabling significant computational efficiency. GDS results under full-parameter are shown in Appendix~\ref{app:full_ft}.

\section{Conclusion}
We revisit pre-training data detection from an optimization and training-dynamics perspective and introduce GDS, a fine-tuning–free method for pre-training membership inference. By analyzing model training dynamics, we show that member and non-member samples exhibit stable and interpretable gradient differences in magnitude, location, and concentration across datasets and architectures. Experiments demonstrate that GDS achieves effective and generalizable performance, surpassing competitive baselines. 

\section*{Limitations}
Our method identifies membership samples relying on inherent gradient discrepancies among instances. However, natural discrepancies stemming from samples of varying reasoning difficulty and different training exposure degrees inevitably interfere with discrimination accuracy, which accordingly degrades the cross-dataset transferability of our approach. In practical scenarios, we only conduct preliminary attempts to integrate supervised and unsupervised inference paradigms, while more effective and in-depth fusion strategies remain to be further explored in future work.

\section*{Impact Statement}
This work investigates pre-training data detection, a form of membership inference that determines whether a given sample was included in a LLM’s pre-training corpus. While our proposed GDS method improves detection accuracy and generalization without fine-tuning, it also raises important ethical concerns. Pre-training data detection methods could be misused to probe proprietary or confidential training corpora, potentially revealing sensitive information about data sources or collection practices. Although GDS does not reconstruct or expose training samples directly, it contributes to capabilities that may be exploited if applied irresponsibly. Our primary motivation is to support transparency, accountability, and safety. As pre-training datasets become increasingly large and opaque, tools like GDS can help researchers, auditors, and regulators verify data usage claims, identify potential copyright violations, and detect benchmark contamination. In this sense, GDS is intended as a diagnostic and auditing mechanism rather than a data extraction attack. We release code and models to promote reproducibility while encouraging responsible deployment. Overall, we hope this work advances trustworthy and transparent large-scale model development while highlighting the need for careful governance of pre-training data.

\bibliography{main}

\appendix

\section{Details of Baseline}
\label{app:baseline_details}

\noindent\textbf{Shared Notations}:
We first define the universal notations used across all baseline methods for consistency:
\begin{itemize}
    \item $x$: Input text sample for membership inference detection.
    \item $\{x_1, x_2, \dots, x_N\}$: Token sequence of input text sample $x$, where $N$ denotes the total number of tokens in the text.
    \item $x_{<t}$: Prefix token sequence of the $t$-th token $x_t$, i.e., $x_{<t} = \{x_1, \dots, x_{t-1}\}$.
    \item $\theta$: Trainable parameters of the pre-trained language model.
    \item $p(\cdot \mid \cdot; \theta)$: Conditional probability distribution parameterized by the model parameters $\theta$.
    \item $k\%$: Percentage threshold for selecting low-probability outlier tokens (used in Min-k\% and Min-k\%++).
\end{itemize}

\subsection{PPL}
Perplexity (PPL) is a classic global likelihood-based metric that measures the model's uncertainty in predicting a given text, with lower PPL indicating higher text familiarity (more likely to be a member sample). First, the text is tokenized into a sequence of tokens $\{x_1, x_2, \dots, x_N\}$. Then, the log-likelihood of each token is calculated conditional on the preceding tokens, and the final PPL is derived by normalizing the sum of log-likelihoods by the token sequence length and performing an exponential transformation.

\begin{equation}
\text{PPL} = \exp\left(-\frac{1}{N} \sum_{i=1}^N \log p(x_i | x_1, ..., x_{i-1}; \theta)\right)
\end{equation}

\subsection{Zlib}
Zlib is a compression-based global likelihood-related method that leverages compression entropy to distinguish member and non-member samples, which calibrates the loss using the input's Zlib entropy. For an input text sample $x$ and a pre-trained model $M$, the Zlib score is defined as the ratio of the loss related to the text and model to the Zlib compression entropy of the input text:

\begin{equation}
\text{Zlib Score} = \frac{L(x, M)}{\text{zlib}(x)}
\label{eq:zlib_score}
\end{equation}

where $L(x, M)$ denotes the loss value associated with the input text $x$ under the model $M$ and
$\text{zlib}(x)$ represents the Zlib compression entropy of the input text $x$, which is derived from the length of the text after Zlib compression.

\subsection{Min-k\%}
Min-k\% is a local likelihood-based method that mitigates global word frequency interference by focusing on low-probability outlier tokens. For an input text token sequence $x = \{x_1, x_2, \dots, x_N\}$ (where $N$ denotes the total number of tokens), it first calculates the conditional log-likelihood of each token $x_i$ given its preceding context:
$$\log p(x_i \mid x_1, x_2, \dots, x_{i-1}; \theta)$$
Subsequently, it sorts all token conditional probabilities in ascending order, selects the bottom $k\%$ tokens (i.e., the $k\%$ tokens with the smallest conditional probabilities) to form the low-probability set $T_{\text{min}(k\%)} = \text{Min-K\%}(x)$. The final Min-k\% score is defined as the average conditional log-likelihood of tokens in $T_{\text{min}(k\%)}$:

\begin{equation}
\begin{gathered}
\text{Min-k\% Score}(x) \\
= \frac{1}{|T_{\text{min}(k\%)}|} \sum_{x_i \in T_{\text{min}(k\%)}} \log p(x_i \mid x_1, \dots, x_{i-1}; \theta)
\label{eq:min_k_score}
\end{gathered}
\end{equation}

where $|T_{\text{min}(k\%)}| = \lfloor k\% \times N \rfloor$ denotes the cardinality of the low-probability token set (rounded down to the nearest integer). During detection, a threshold is set on $\text{Min-k\% Score}(x)$: text with a score below the threshold is classified as a member sample (since member samples tend to contain more low-probability outlier tokens), and vice versa.

\subsection{Min-k\%++}
Min-k\%++ is an optimized variant of Min-k\% that enhances detection robustness by integrating vocabulary-level probability statistics, addressing the text-length dependency and short-text fluctuation issues of the original method. For an input text token sequence $x = \{x_1, x_2, \dots, x_N\}$, it first computes the normalized conditional log-likelihood for each token $x_t$ (given prefix $x_{<t} = \{x_1, \dots, x_{t-1}\}$) by normalizing with the vocabulary-wide log-probability statistics of the prefix $x_{<t}$.

\textbf{Core Token-Level Score Calculation}

For each token $x_t$, the token-level Min-k\%++ score is defined as:
\begin{equation}
\begin{gathered}
\text{Min-k\%++}_{\text{token}}(x_{<t}, x_t) \\
= \frac{\log p(x_t \mid x_{<t}; \theta) - \mu_{x_{<t}}}{\sigma_{x_{<t}}}
\label{eq:min_kpp_token}
\end{gathered}
\end{equation}
where $\mu_{x_{<t}}=\mathbb{E}_{z \sim p(\cdot \mid x_{<t}; \theta)} \left[ \log p(z \mid x_{<t}; \theta) \right]$: Expectation of the next-token log-likelihood over the model's vocabulary, given prefix $x_{<t}$; $\sigma_{x_{<t}}=\sqrt{\mathbb{E}_{z \sim p(\cdot \mid x_{<t}; \theta)} \left[ \left( \log p(z \mid x_{<t}; \theta) - \mu_{x_{<t}} \right)^2 \right]}$: Standard deviation of the next-token log-likelihood over the vocabulary, given prefix $x_{<t}$.

Both $\mu_{x_{<t}}$ and $\sigma_{x_{<t}}$ can be computed analytically using the model's output logits (since $p(\cdot \mid x_{<t}; \theta)$ follows a categorical distribution), requiring no additional computational overhead beyond standard LLM inference.

\textbf{Final Sequence-Level Score Calculation}

Similar to Min-k\%, Min-k\%++ selects the bottom $k\%$ tokens with the smallest $\text{Min-k\%++}_{\text{token}}$ scores to form the low-probability set $T_{\text{min}(k\%)}^+$. The final sequence-level score is the average of the token-level scores in this set:
\begin{equation}
\begin{gathered}
\text{Min-k\%++ Score}(x) \\
= \frac{1}{|T_{\text{min}(k\%)}^+|} \sum_{(x_{<t}, x_t) \in T_{\text{min}(k\%)}^+} \text{Min-k\%++}_{\text{token}}(x_{<t}, x_t)
\label{eq:min_kpp_score}
\end{gathered}
\end{equation}

where $|T_{\text{min}(k\%)}^+| = \lfloor k\% \times N \rfloor$ denotes the cardinality of the low-probability token set. During detection, a threshold is applied: text with a lower Min-k\%++ score is more likely to be a member sample, as the normalized score better distinguishes low-probability outlier tokens of member samples from non-member ones.

\subsection{FSD}
Fine-tuned Score Deviation (FSD) is a fine-tuning enhanced method that leverages score discrepancies between pre-trained and fine-tuned models for member sample detection. Its core workflow includes two key steps: first, constructing a domain-matched non-member fine-tuning dataset (using post-model-release unseen data), then performing self-supervised next-token prediction fine-tuning on the pre-trained model. Finally, it calculates the score difference between the pre-trained and fine-tuned models to distinguish member and non-member samples. The self-supervised fine-tuning loss is defined as:
\begin{equation}
\mathcal{L}_{\text{fine-tuning}}(x) = -\frac{1}{n} \sum_{i=1}^n \log f_\theta(x_i \mid x_1, \dots, x_{i-1})
\label{eq:fsd_finetune_loss}
\end{equation}

The final FSD score is the deviation between the sample scores obtained from the pre-trained model $f_\theta$ and the fine-tuned model $f_{\theta'}$ (with parameters $\theta'$):
\begin{equation}
\text{FSD}(x; f_\theta, f_{\theta'}) = S(x; f_\theta) - S(x; f_{\theta'})
\label{eq:fsd_score}
\end{equation}

where $S(\cdot)$ represents a base scoring function (e.g., Perplexity, Min-k\%). A large FSD score indicates the sample is likely a non-member (fine-tuning reduces non-member perplexity significantly), while a small score suggests a member sample (fine-tuning has negligible impact on member perplexity). A validation-set optimized threshold $\varepsilon$ is used for final classification.

\subsection{Experimental Settings}
\label{app:experiment_setting}
We use the PEFT library~\cite{mangrulkar2022peft} for LoRA-based~\cite{hu2022lora} gradient feature extraction without parameter updates. The base model is initialized from pre-trained checkpoints, and for LoRA configuration, we adopt default settings with rank $r=16$, LoRA scaling factor $\alpha=32$, and dropout set to 0 since we only focus on gradients from a single backpropagation step and no parameter updates are required.
Target modules include all submodules of attention and FFN, including query, key, value, output, gate, up, and down projections. For different datasets, we select the tokenizer sequence length based on the overall length distribution: 256 for the WikiMIA dataset and 512 for all other datasets. When training the MLP classifier, the dataset is uniformly split into training and validation sets at a ratio of 3:7. The MLP classifier is configured with two hidden layers of dimensions 128 and 64, respectively, and uses the default Adam optimizer with a learning rate of 0.001. An early stopping strategy is implemented to avoid overfitting.

\section{Motivation Experiment Supplementary Details}
\label{app:motivation_experiment}

To verify the parameter update law driven by the training process, we conduct LoRA fine-tuning with 8 epochs on LLaMA-7B using the unseen set of BookMIA dataset, counting LoRA matrix parameter updates per epoch to extract dynamic change curves of target features. Experiments are performed under three learning rates (\ensuremath{1e-5}, \ensuremath{3e-5}, \ensuremath{5e-5}). We present the variation trends of the four feature categories across different modules and layers under each learning rate.

\begin{figure*}[htbp]
    \centering
    \begin{subfigure}{0.23\textwidth}
        \centering
        \includegraphics[width=\textwidth]{pictures/test_final_b1.pdf}
        \caption{Module-Mean Update}
        \label{subfig:eta1_mod_mean}
    \end{subfigure}
    \hfill
    \begin{subfigure}{0.23\textwidth}
        \centering
        \includegraphics[width=\textwidth]{pictures/test_final_b3.pdf}
        \caption{Module-Sparsity}
        \label{subfig:eta1_mod_sparsity}
    \end{subfigure}
    \hfill
    \begin{subfigure}{0.23\textwidth}
        \centering
        \includegraphics[width=\textwidth]{pictures/test_final_b2.pdf}
        \caption{Module-Top10\% Ratio}
        \label{subfig:eta1_mod_top10}
    \end{subfigure}
    \hfill
    \begin{subfigure}{0.23\textwidth}
        \centering
        \includegraphics[width=\textwidth]{pictures/test_final_b4.pdf}
        \caption{Module-Row Eccentricity}
        \label{subfig:eta1_mod_row_ecc}
    \end{subfigure}

    \vspace{1em}
    \begin{subfigure}{0.23\textwidth}
        \centering
        \includegraphics[width=\textwidth]{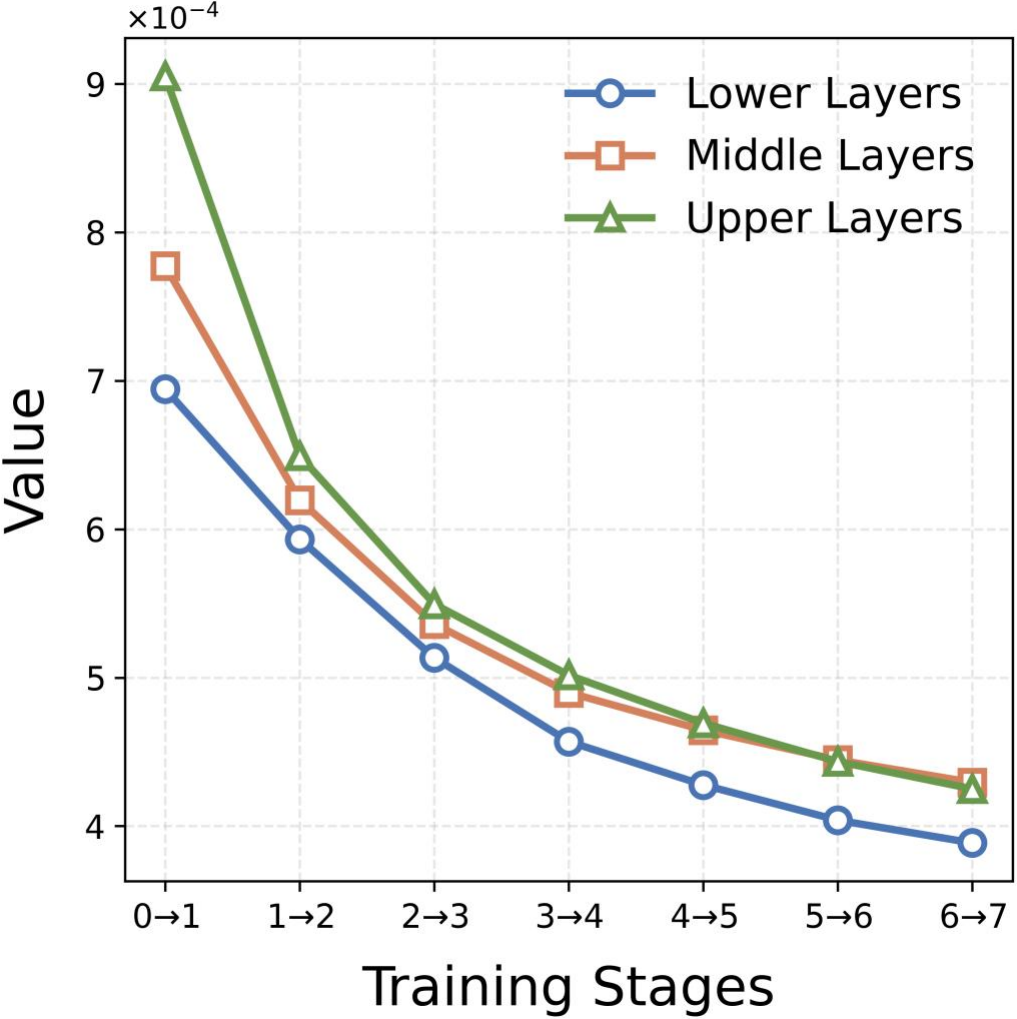}
        \caption{Layer-Mean Update}
        \label{subfig:eta1_layer_mean}
    \end{subfigure}
    \hfill
    \begin{subfigure}{0.23\textwidth}
        \centering
        \includegraphics[width=\textwidth]{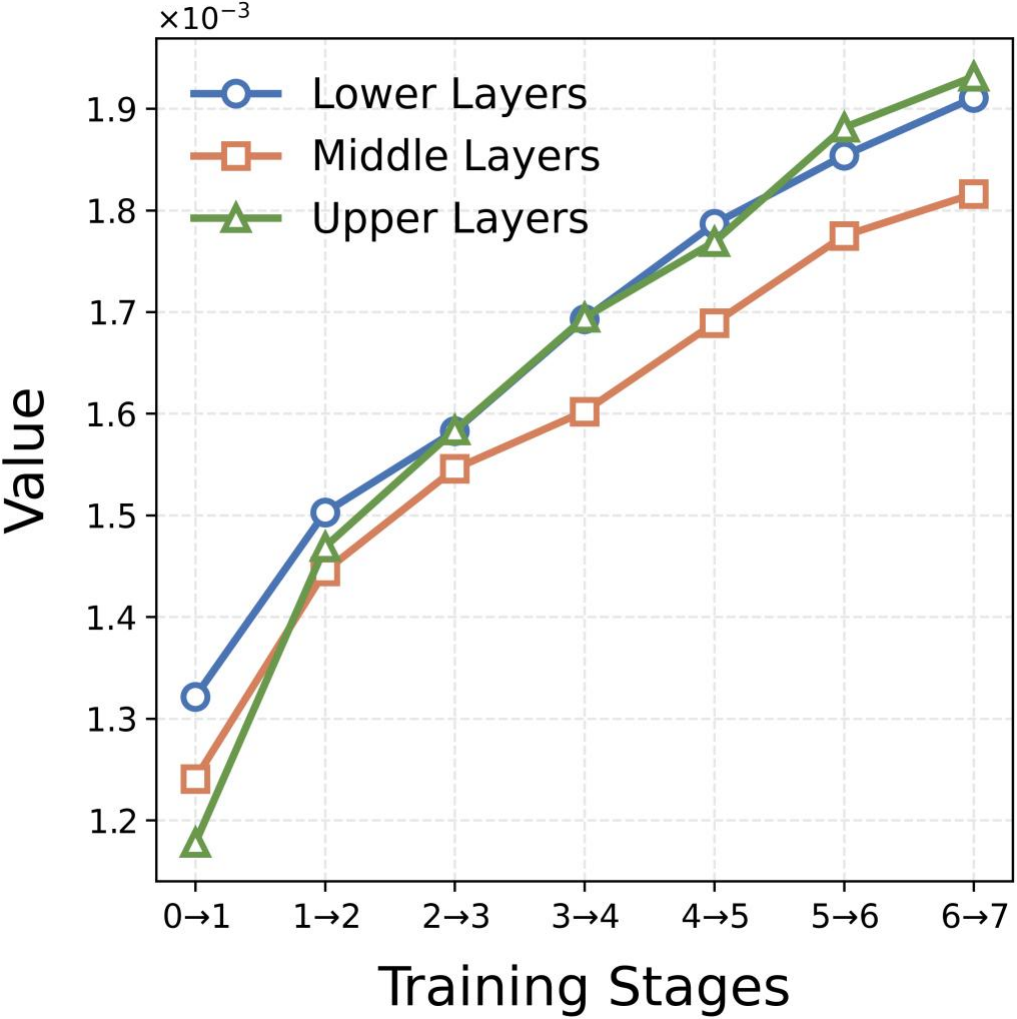}
        \caption{Layer-Sparsity}
        \label{subfig:eta1_layer_sparsity}
    \end{subfigure}
    \hfill
    \begin{subfigure}{0.23\textwidth}
        \centering
        \includegraphics[width=\textwidth]{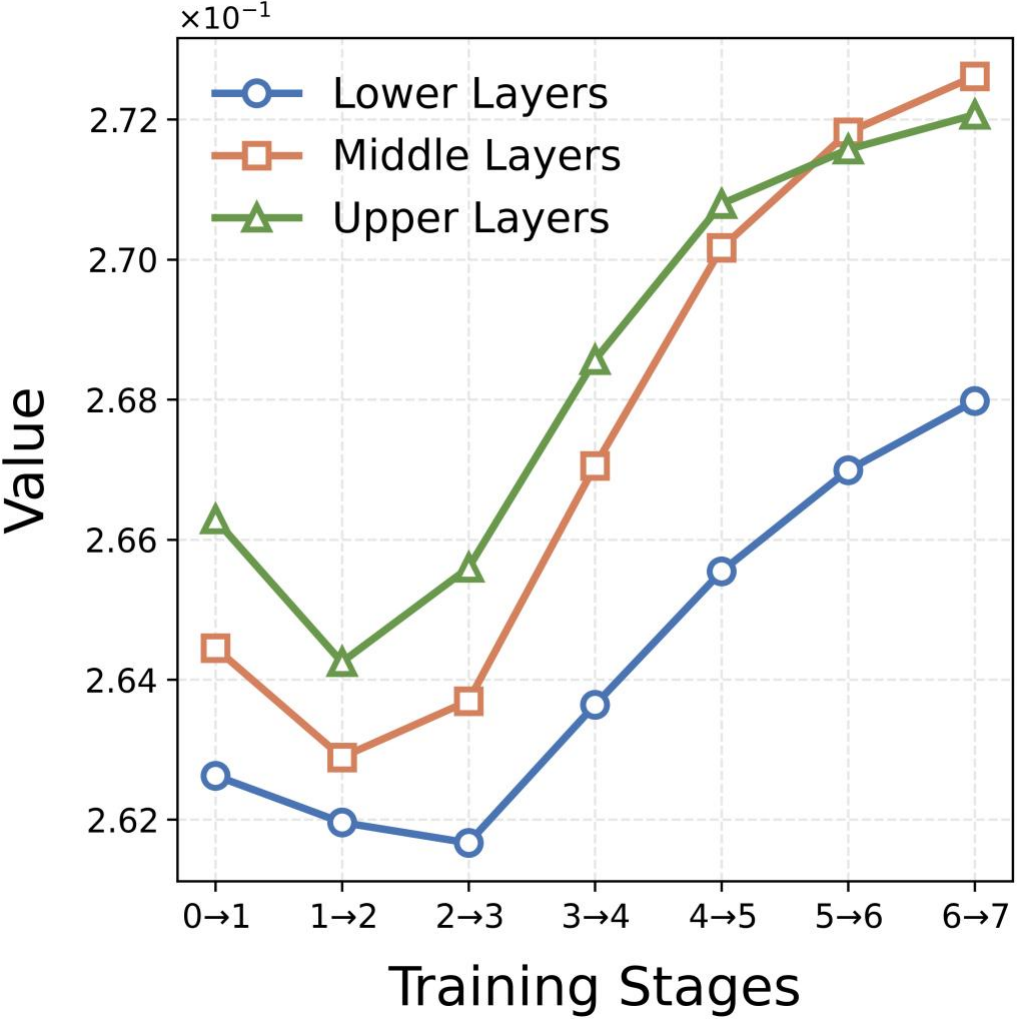}
        \caption{Layer-Top10\% Ratio}
        \label{subfig:eta1_layer_top10}
    \end{subfigure}
    \hfill
    \begin{subfigure}{0.23\textwidth}
        \centering
        \includegraphics[width=\textwidth]{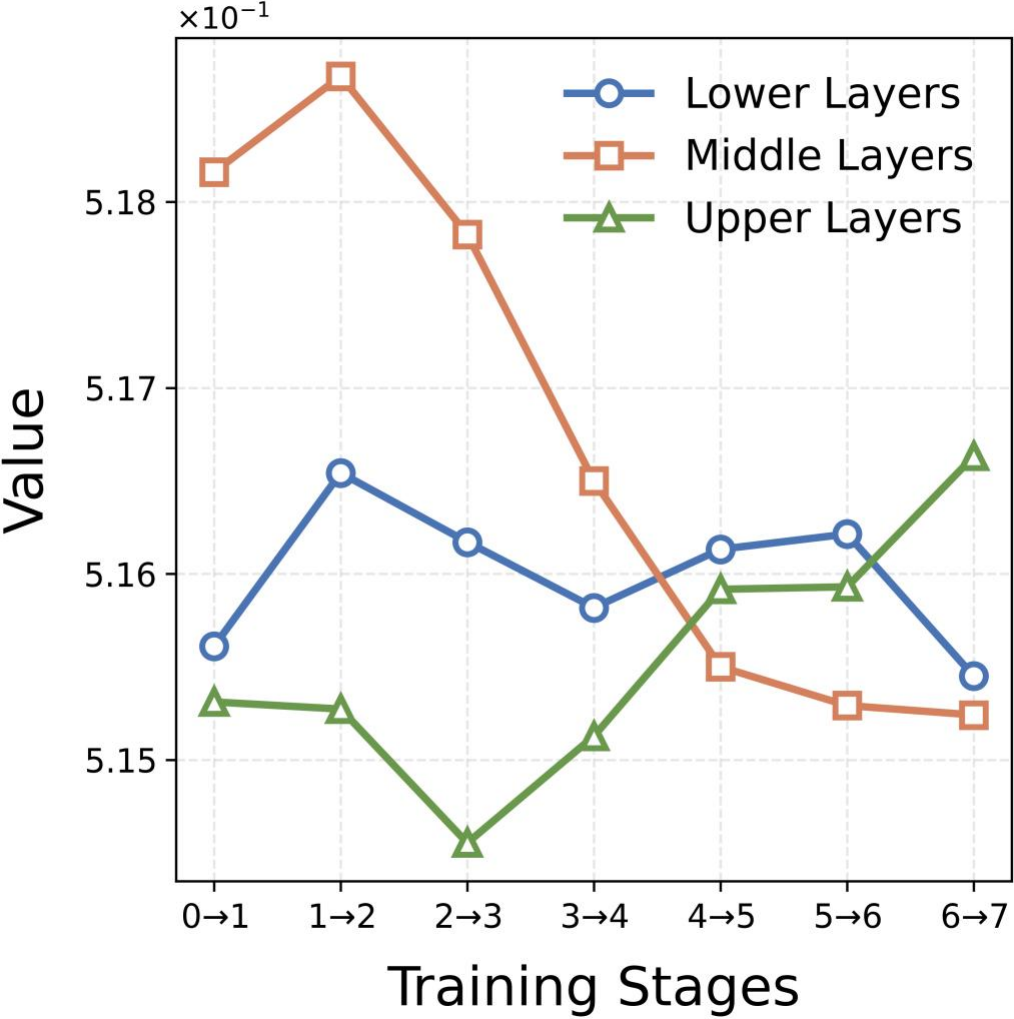}
        \caption{Layer-Col Eccentricity}
        \label{subfig:eta1_layer_col_ecc}
    \end{subfigure}

    \caption{Feature Dynamic Changes During LoRA Fine-Tuning (Learning Rate \ensuremath{1e-5}). Top row: Features in ATT/MLP modules; Bottom row: Features in low/middle/high layers.}
    \label{fig:motivation_eta1}
\end{figure*}

\begin{figure*}[htbp]
    \centering
    \begin{subfigure}{0.23\textwidth}
        \centering
        \includegraphics[width=\textwidth]{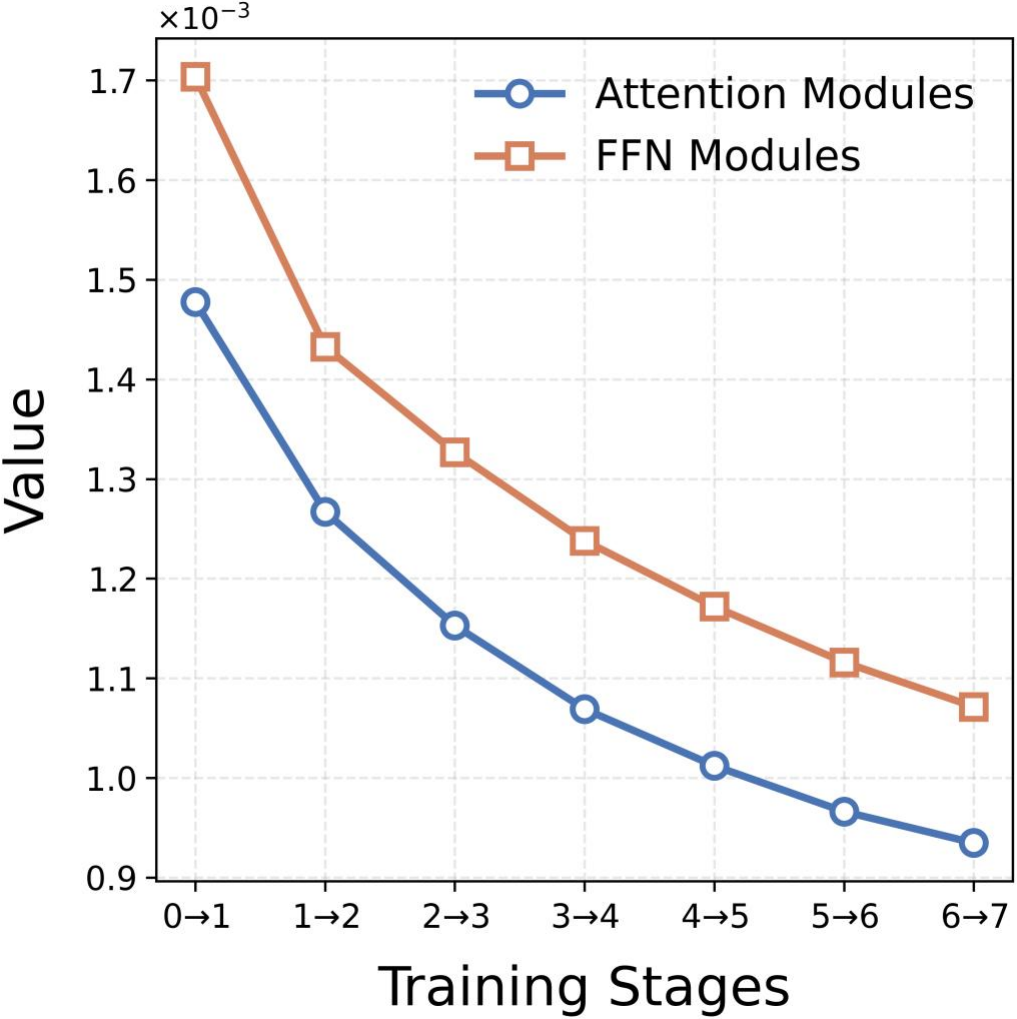}
        \caption{Module-Mean Update}
        \label{subfig:eta2_mod_mean}
    \end{subfigure}
    \hfill
    \begin{subfigure}{0.23\textwidth}
        \centering
        \includegraphics[width=\textwidth]{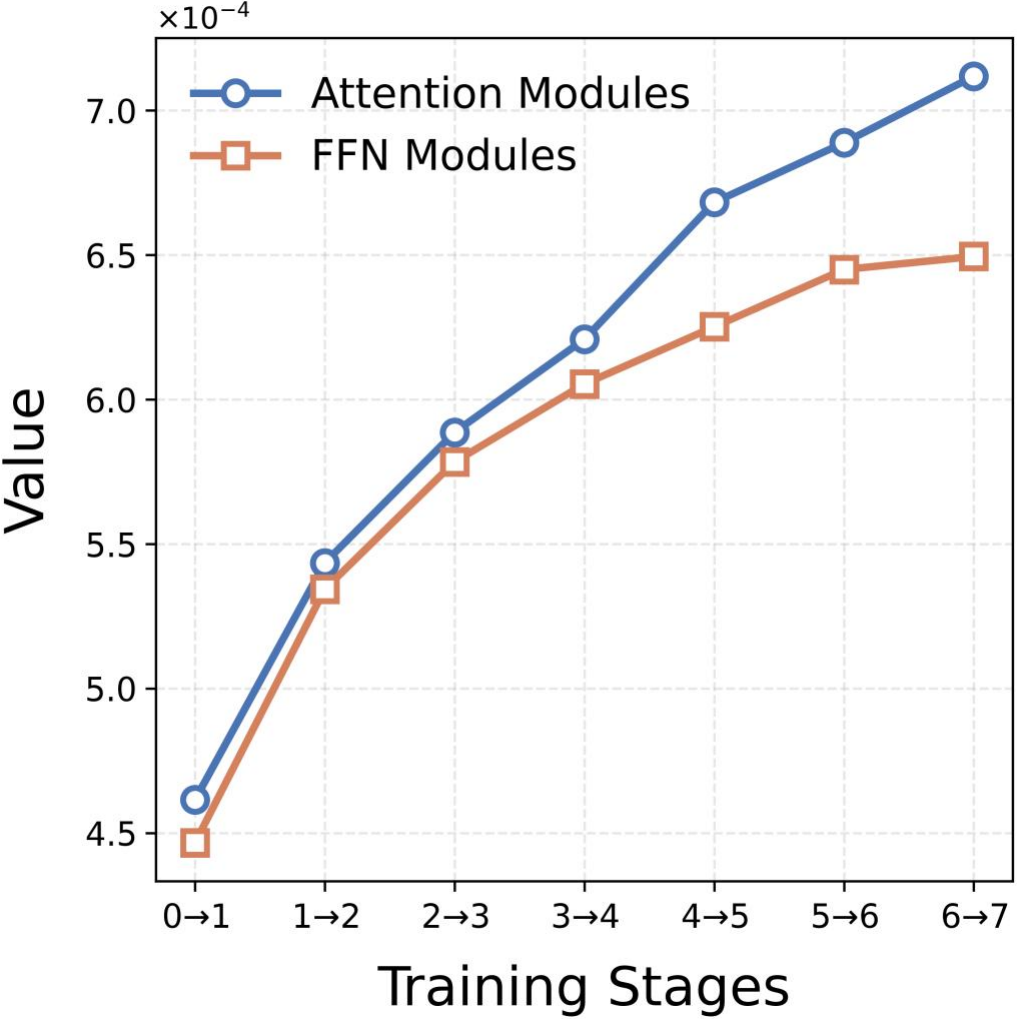}
        \caption{Module-Sparsity}
        \label{subfig:eta2_mod_sparsity}
    \end{subfigure}
    \hfill
    \begin{subfigure}{0.23\textwidth}
        \centering
        \includegraphics[width=\textwidth]{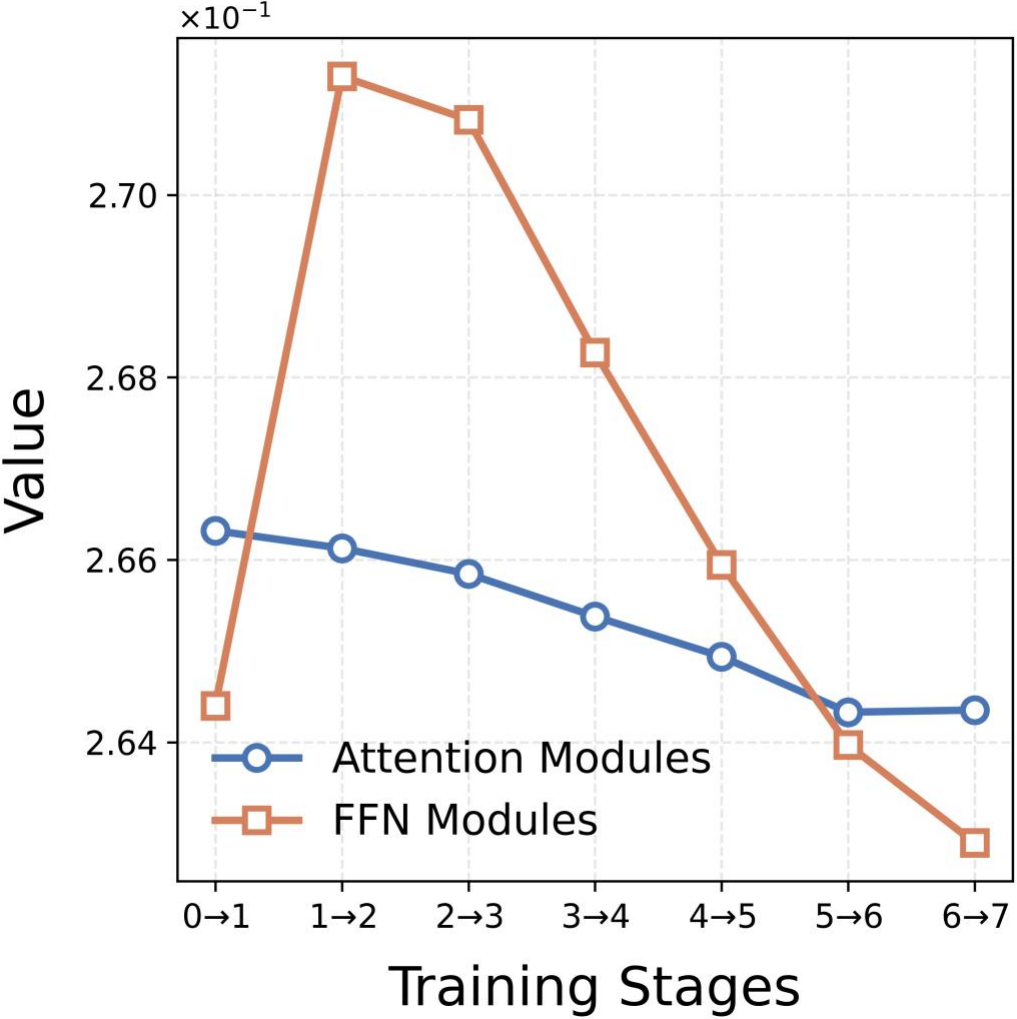}
        \caption{Module-Top10\% Ratio}
        \label{subfig:eta2_mod_top10}
    \end{subfigure}
    \hfill
    \begin{subfigure}{0.23\textwidth}
        \centering
        \includegraphics[width=\textwidth]{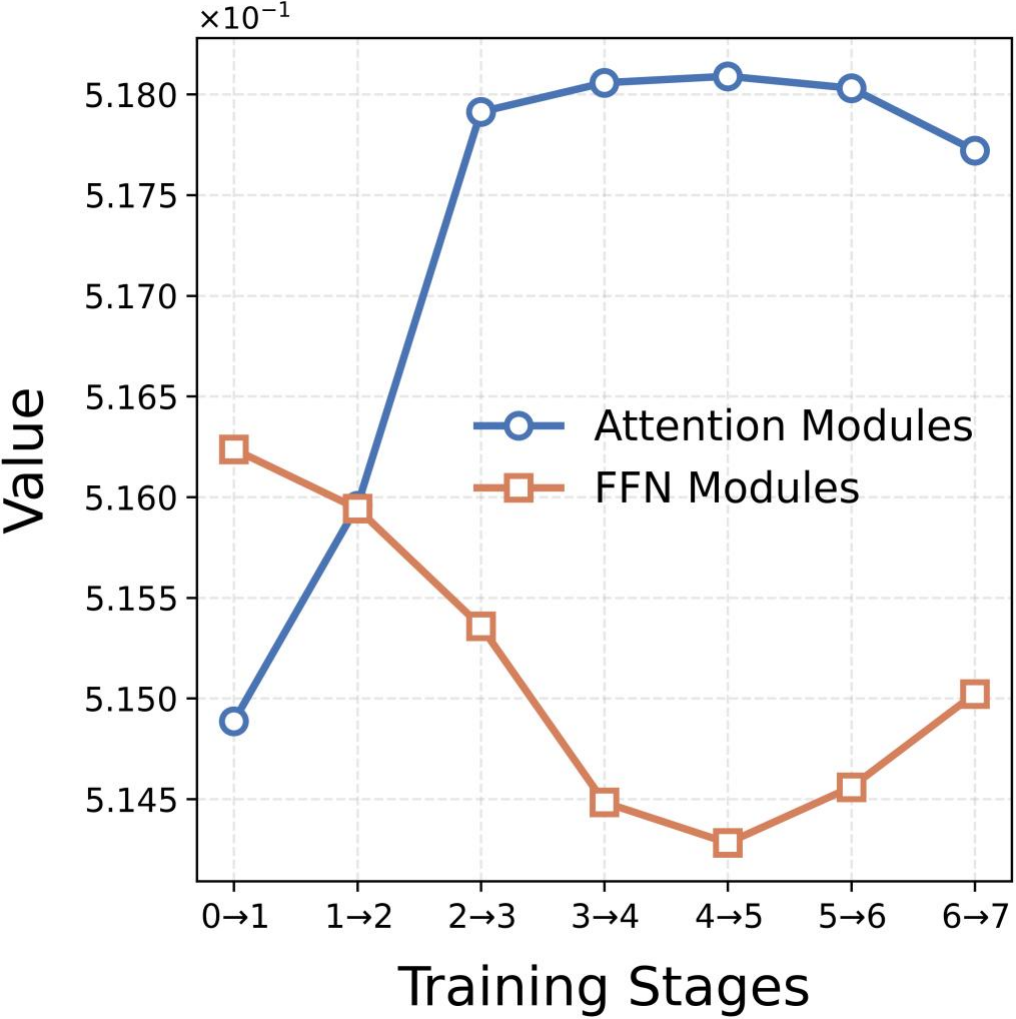}
        \caption{Module-Row Eccentricity}
        \label{subfig:eta2_mod_row_ecc}
    \end{subfigure}

    \vspace{1em}
    \begin{subfigure}{0.23\textwidth}
        \centering
        \includegraphics[width=\textwidth]{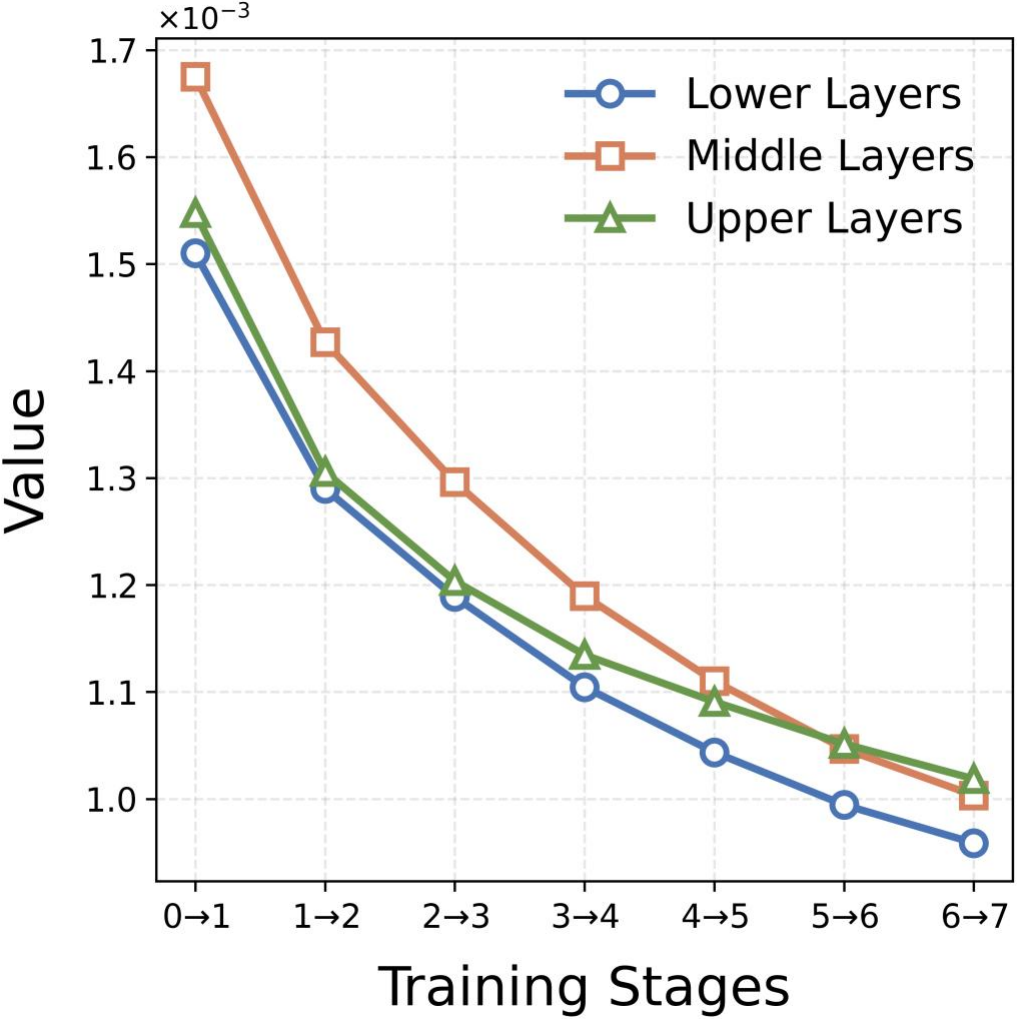}
        \caption{Layer-Mean Update}
        \label{subfig:eta2_layer_mean}
    \end{subfigure}
    \hfill
    \begin{subfigure}{0.23\textwidth}
        \centering
        \includegraphics[width=\textwidth]{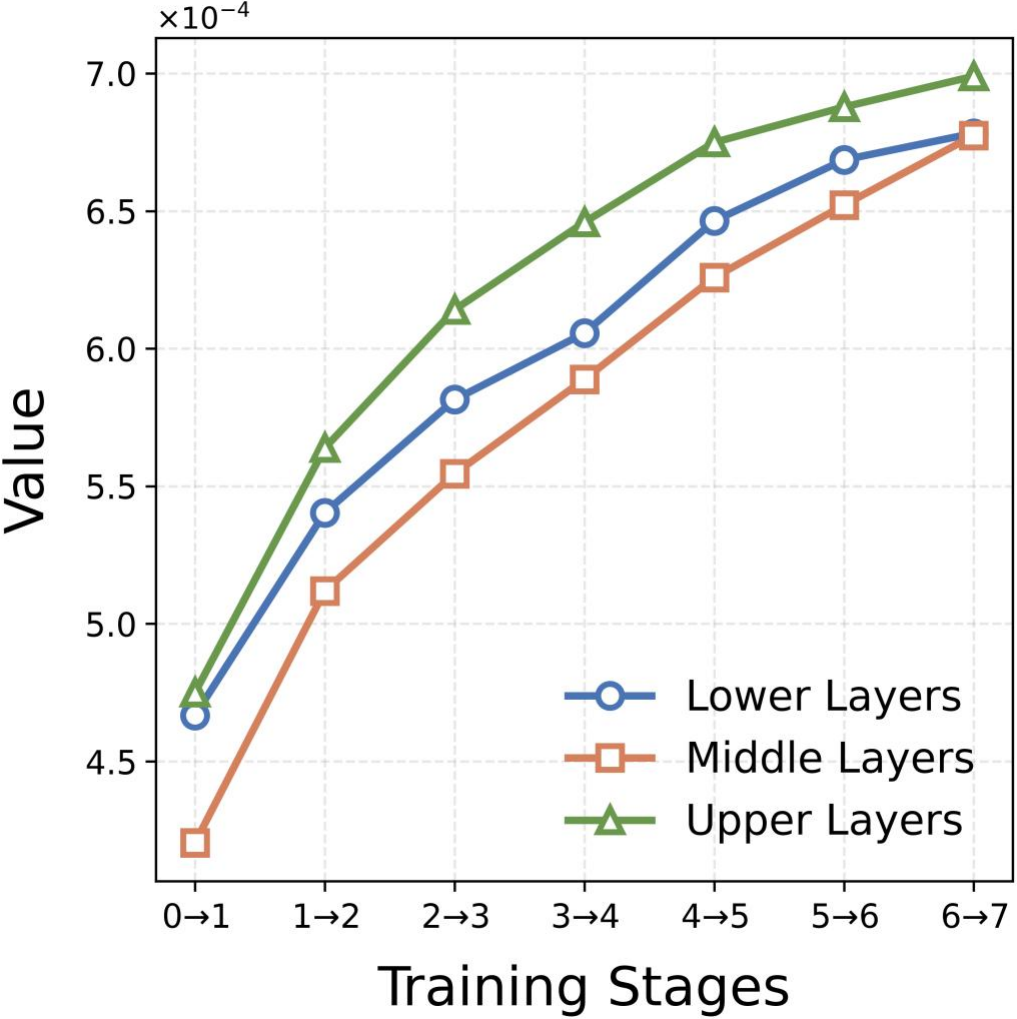}
        \caption{Layer-Sparsity}
        \label{subfig:eta2_layer_sparsity}
    \end{subfigure}
    \hfill
    \begin{subfigure}{0.23\textwidth}
        \centering
        \includegraphics[width=\textwidth]{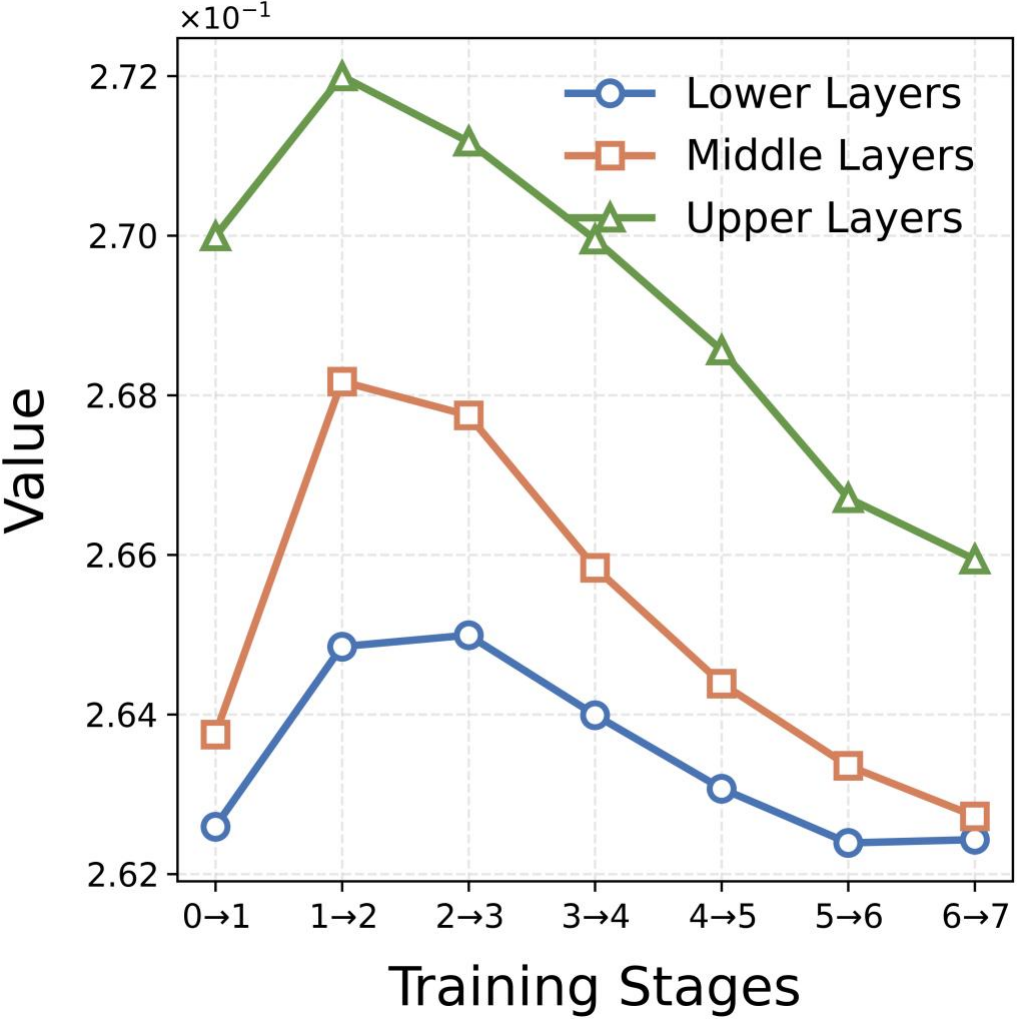}
        \caption{Layer-Top10\% Ratio}
        \label{subfig:eta2_layer_top10}
    \end{subfigure}
    \hfill
    \begin{subfigure}{0.23\textwidth}
        \centering
        \includegraphics[width=\textwidth]{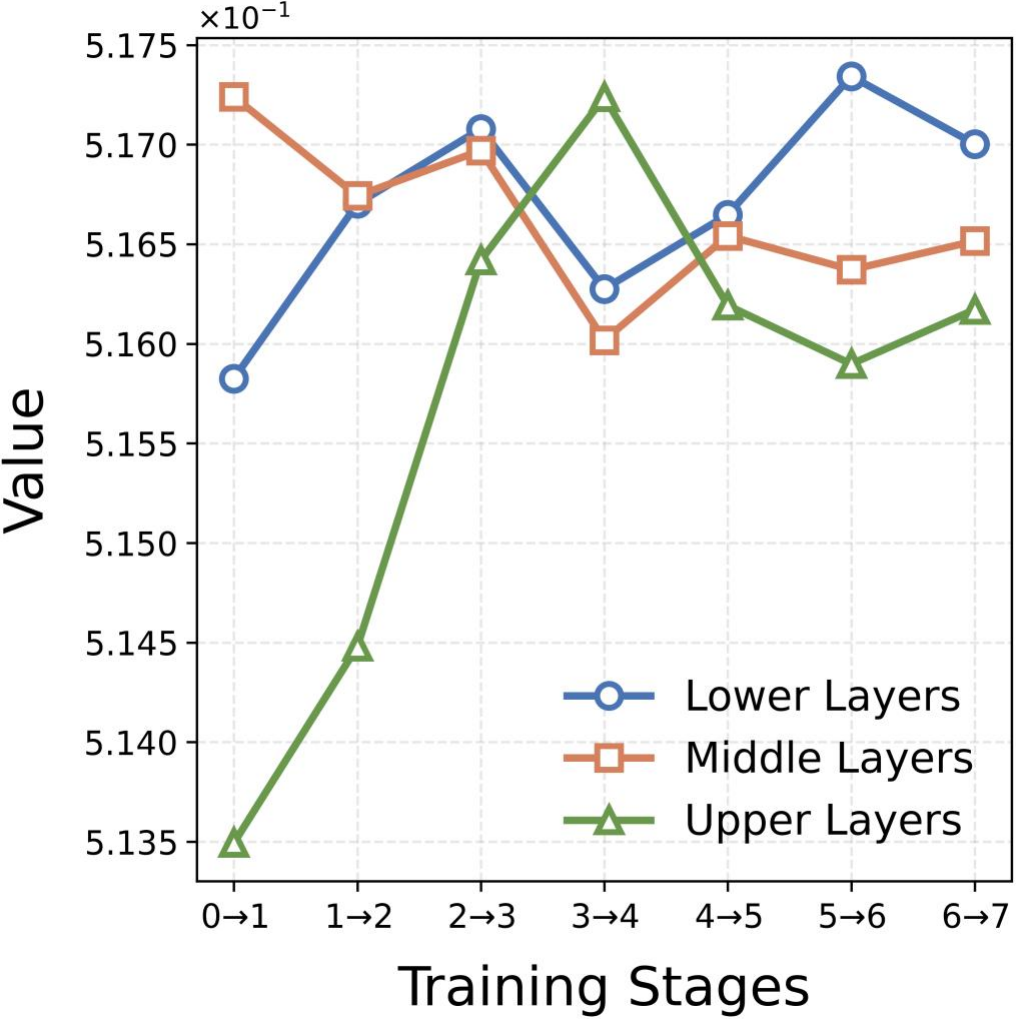}
        \caption{Layer-Col Eccentricity}
        \label{subfig:eta2_layer_col_ecc}
    \end{subfigure}

    \caption{Feature Dynamic Changes During LoRA Fine-Tuning (Learning Rate \ensuremath{3e-5}).}
    \label{fig:motivation_eta2}
\end{figure*}

\begin{figure*}[htbp]
    \centering
    \begin{subfigure}{0.23\textwidth}
        \centering
        \includegraphics[width=\textwidth]{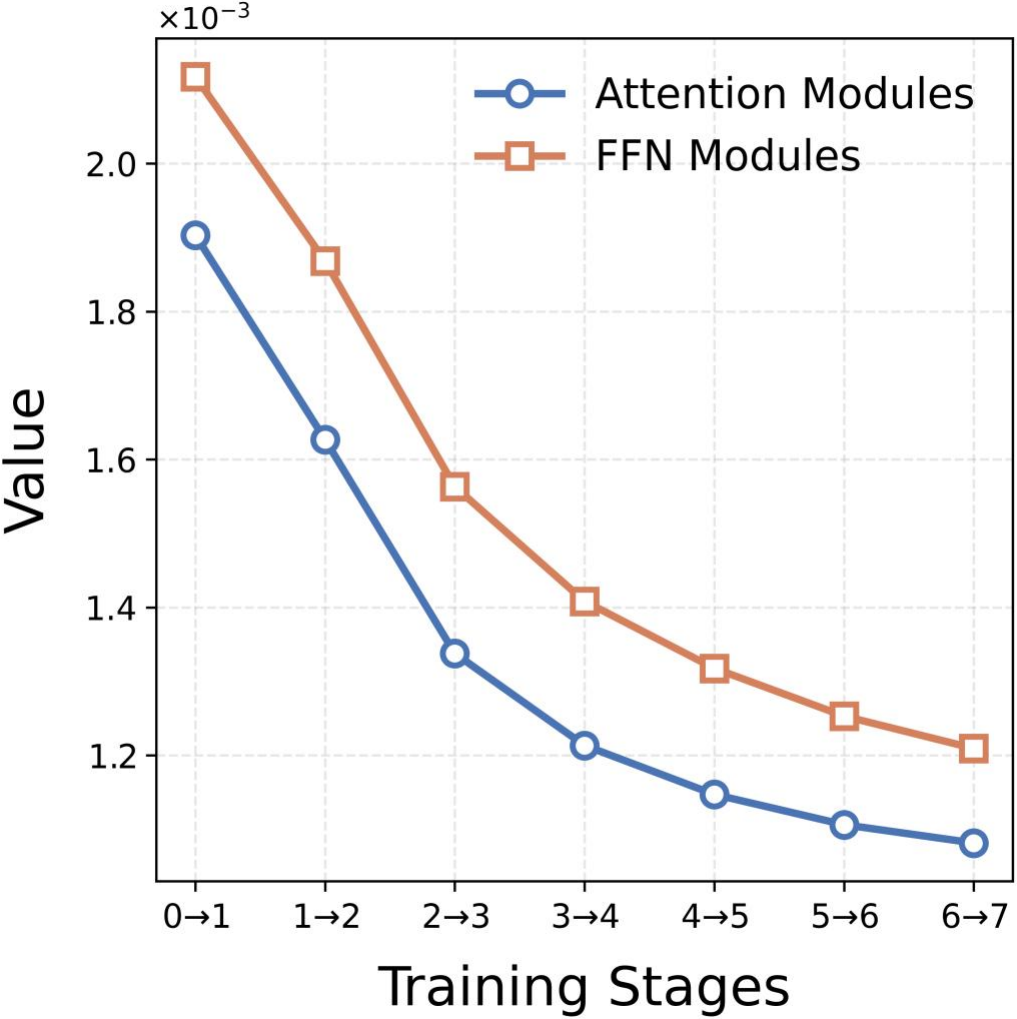}
        \caption{Module-Mean Update}
        \label{subfig:eta3_mod_mean}
    \end{subfigure}
    \hfill
    \begin{subfigure}{0.23\textwidth}
        \centering
        \includegraphics[width=\textwidth]{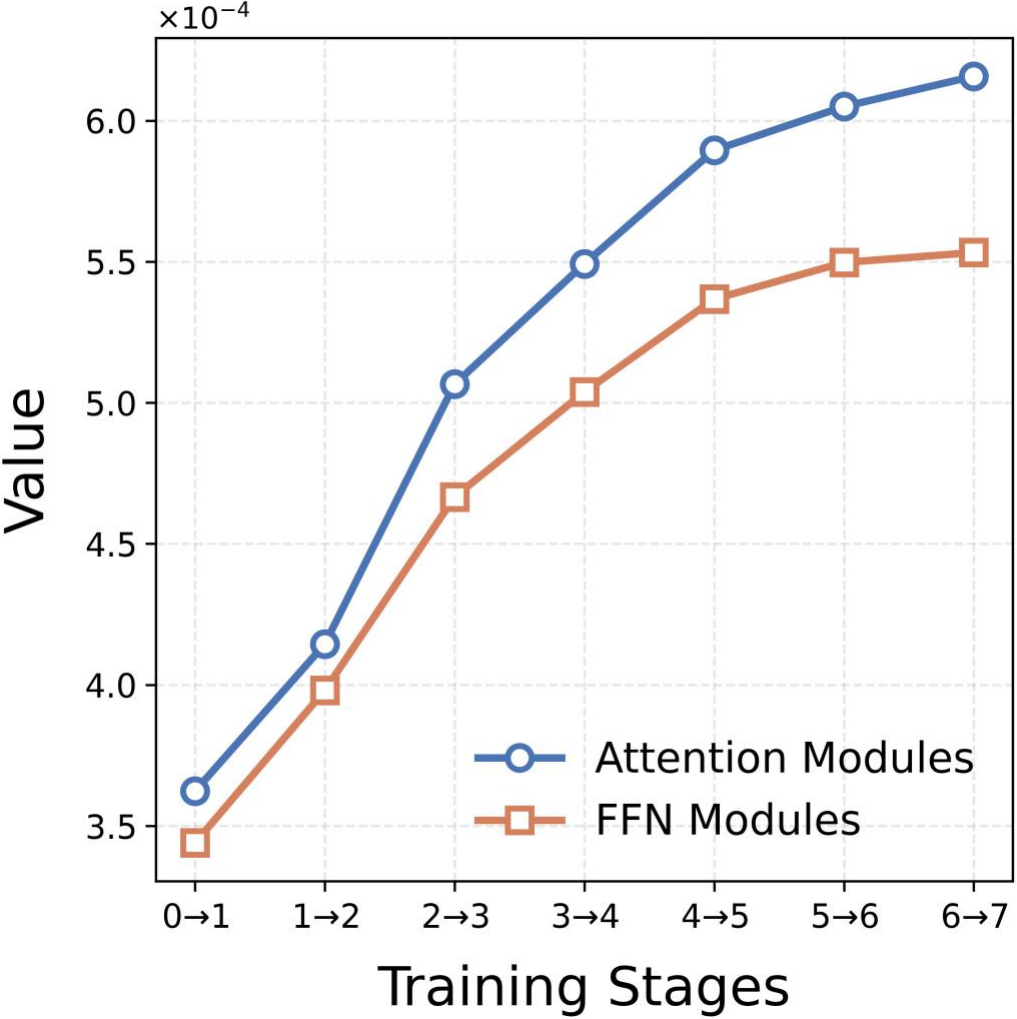}
        \caption{Module-Sparsity}
        \label{subfig:eta3_mod_sparsity}
    \end{subfigure}
    \hfill
    \begin{subfigure}{0.23\textwidth}
        \centering
        \includegraphics[width=\textwidth]{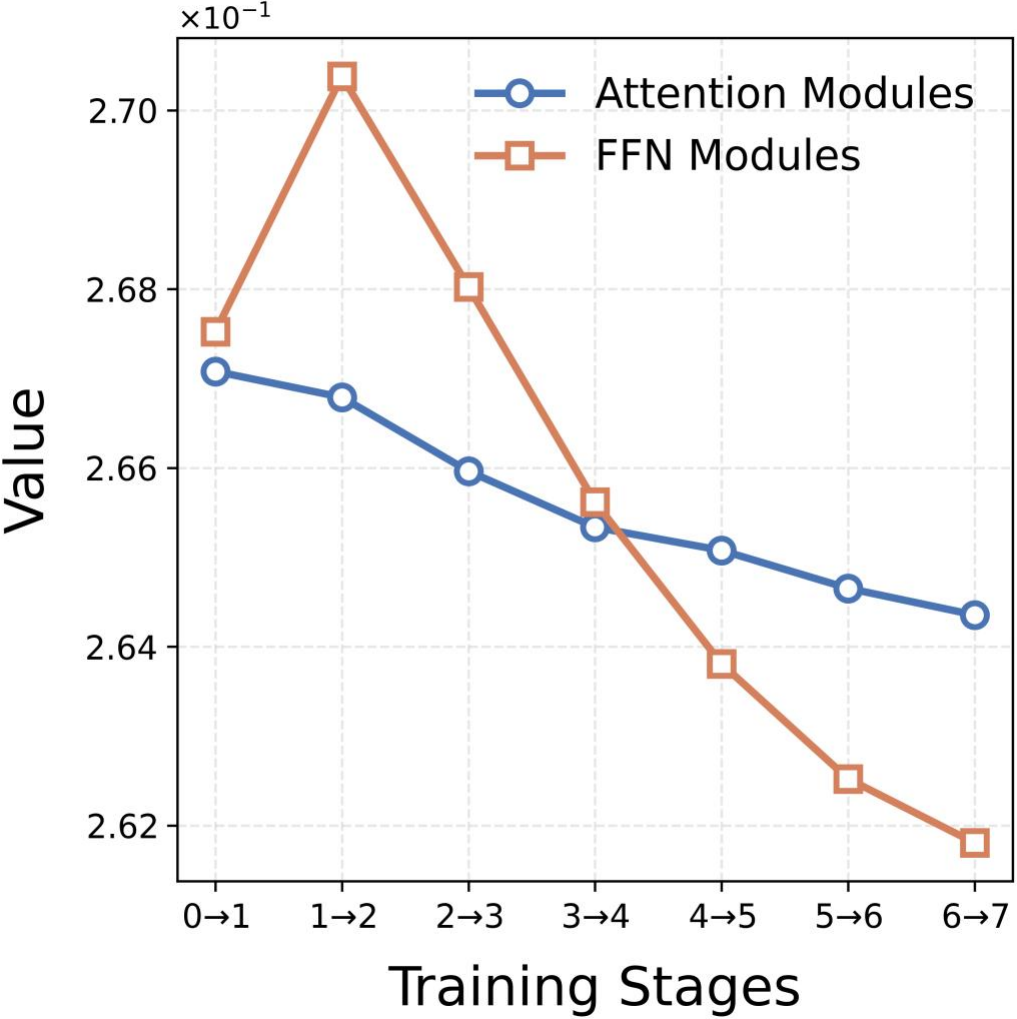}
        \caption{Module-Top10\% Ratio}
        \label{subfig:eta3_mod_top10}
    \end{subfigure}
    \hfill
    \begin{subfigure}{0.23\textwidth}
        \centering
        \includegraphics[width=\textwidth]{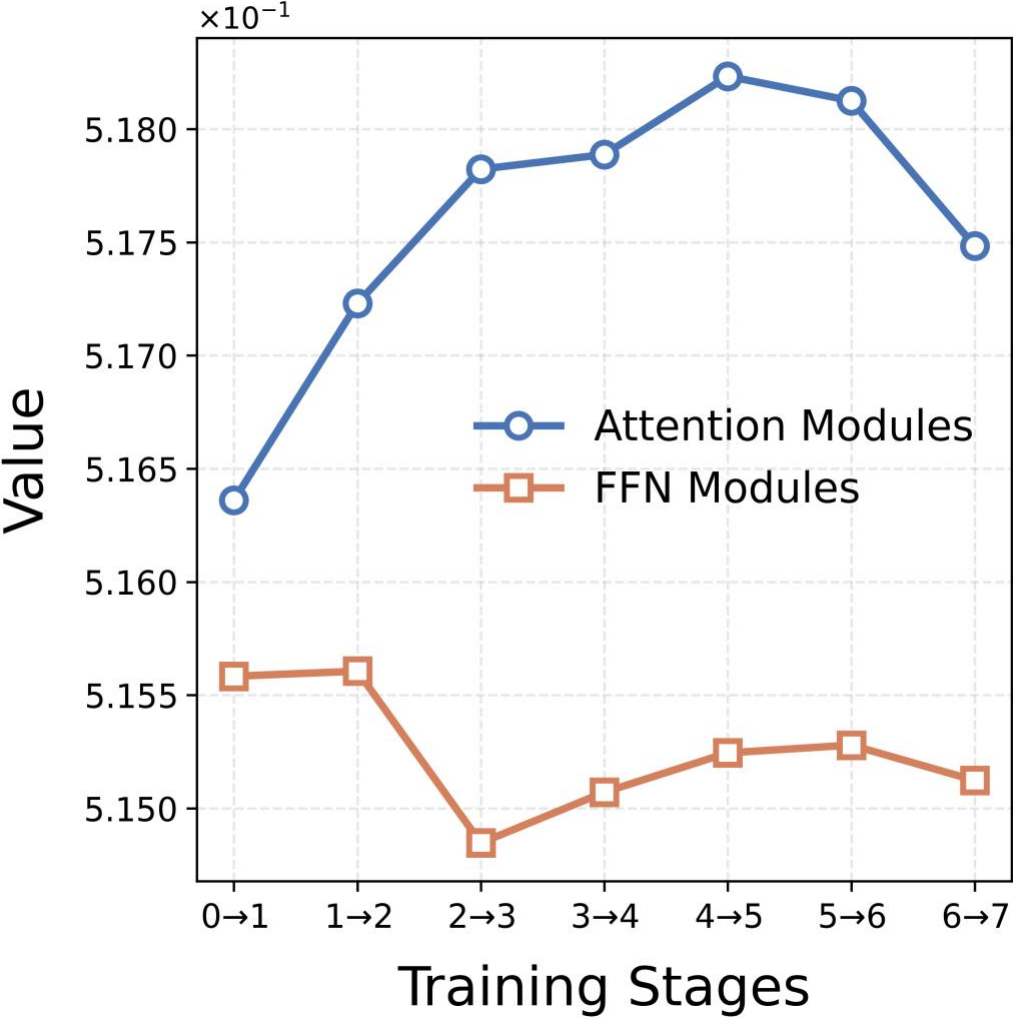}
        \caption{Module-Row Eccentricity}
        \label{subfig:eta3_mod_row_ecc}
    \end{subfigure}

    \vspace{1em}
    \begin{subfigure}{0.23\textwidth}
        \centering
        \includegraphics[width=\textwidth]{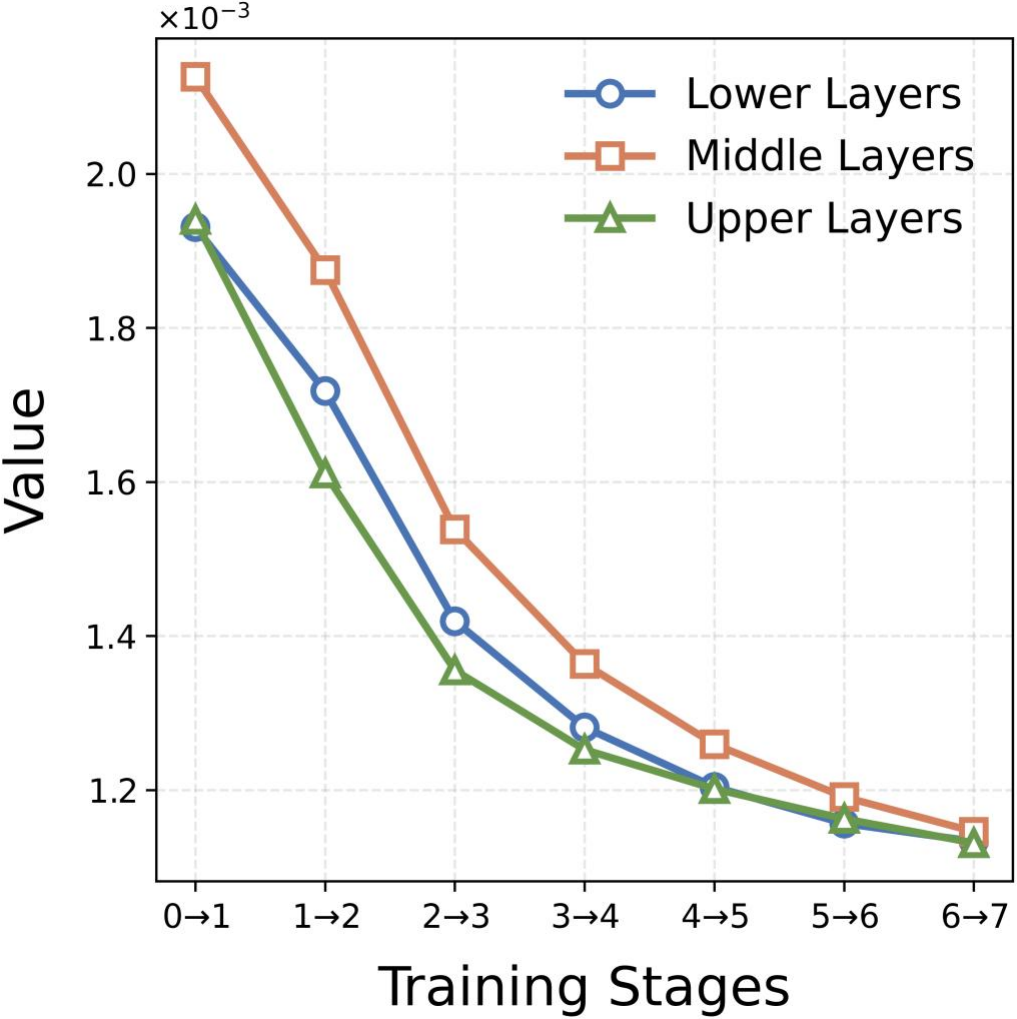}
        \caption{Layer-Mean Update}
        \label{subfig:eta3_layer_mean}
    \end{subfigure}
    \hfill
    \begin{subfigure}{0.23\textwidth}
        \centering
        \includegraphics[width=\textwidth]{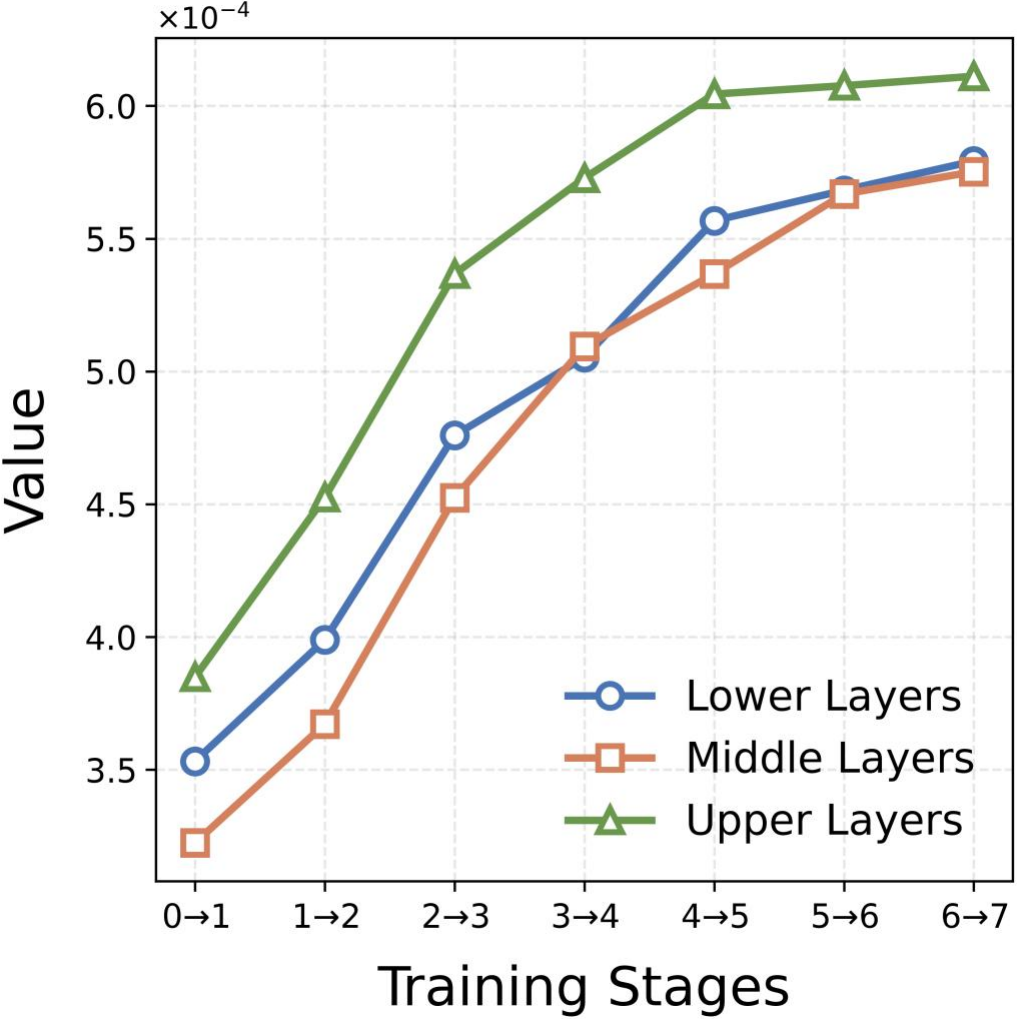}
        \caption{Layer-Sparsity}
        \label{subfig:eta3_layer_sparsity}
    \end{subfigure}
    \hfill
    \begin{subfigure}{0.23\textwidth}
        \centering
        \includegraphics[width=\textwidth]{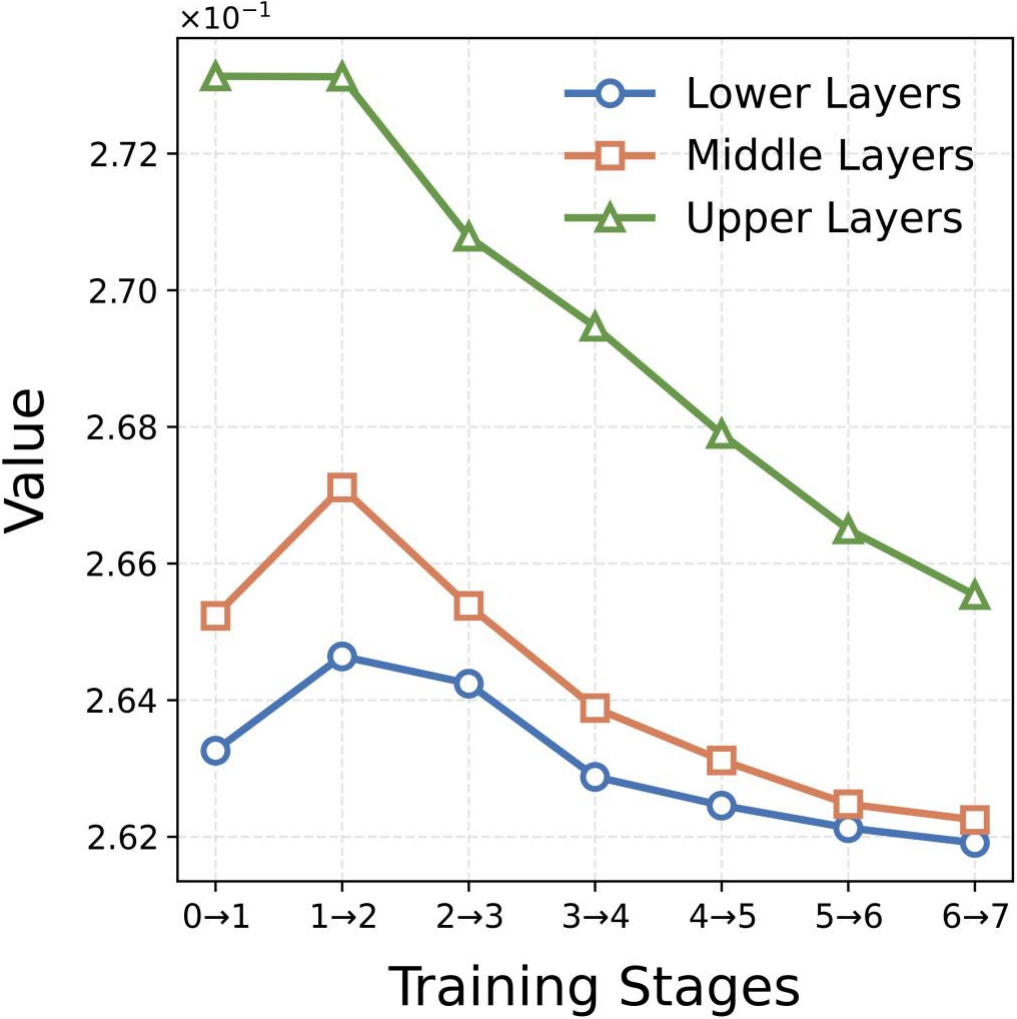}
        \caption{Layer-Top10\% Ratio}
        \label{subfig:eta3_layer_top10}
    \end{subfigure}
    \hfill
    \begin{subfigure}{0.23\textwidth}
        \centering
        \includegraphics[width=\textwidth]{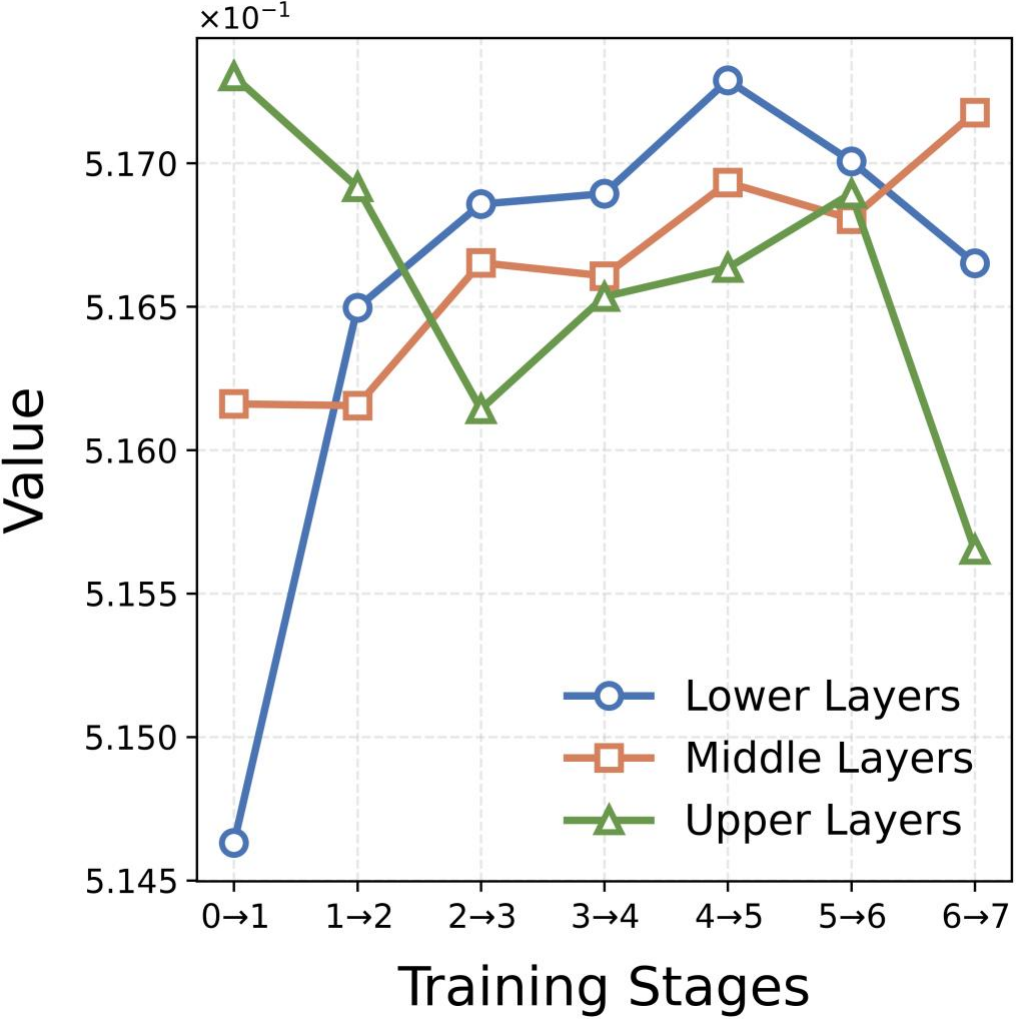}
        \caption{Layer-Col Eccentricity}
        \label{subfig:eta3_layer_col_ecc}
    \end{subfigure}

    \caption{Feature Dynamic Changes During LoRA Fine-Tuning (Learning Rate \ensuremath{5e-5}).}
    \label{fig:motivation_eta3}
\end{figure*}

We present the variation trends of the four feature categories across different modules and layers under each learning rate (Figs.~\ref{fig:motivation_eta1}--\ref{fig:motivation_eta3}). Across all learning rates, $\Delta\Theta_t$ and $S_t$ exhibit clear epoch-wise patterns: $\Delta\Theta_t$ continuously decreases with training epochs, while $S_t$ gradually increases. For $E_t$, distinct opposite trends are observed between the ATT and MLP modules—the update positions of the ATT module tend to be centralized, whereas those of the MLP module tend to be marginalized. Despite this divergence, $E_t$ values of both modules converge to the same stable range across the three learning rates. In contrast, $T_{10,t}$ shows a different dynamic: under large learning rates, it first increases and then decreases, yet remains stably around 0.27 throughout training. Notably, the inflection point of $T_{10,t}$ decline coincides with the stage when $E_t$ stabilizes. This aligns with our core hypothesis: during training, the model identifies core parameters correlated with the input data and performs concentrated updates on them; once this process is completed, the overall parameter update dynamics tend to stabilize. Regarding different hierarchical levels, only $E_t$ shows noticeable discrepancies, while all other features follow the same evolutionary laws.

\section{Ablation Experiments on Sub-features and Sub-modules}
\label{app:ablation_sub}

To further explore the discriminative power of fine-grained components in our method, we conduct two additional ablation experiments: sub-feature category ablation (8 sub-features) and sub-module ablation (7 sub-modules). All experiments follow the same configuration as the main text (LLaMA-7B model, WikiMIA/ArxivTection datasets), with AUC and TPR@5\% FPR as evaluation metrics.

\subsection{Sub-feature Category Ablation Experiment}
The four core feature categories are decomposed into 8 sub-features, as follows:
\begin{itemize}
    \item Magnitude Distribution Features: Abs\_Mean, Std
    \item Core Contribution Features: 10p\_Ratio, Sparsity
    \item Position Offset Features: Row\_Ecc, Col\_Ecc
    \item Row-Dimension Consistency Features: Row\_Mean\_Max, Row\_Mean\_Std
\end{itemize}

Ablation is performed by removing one sub-feature at a time (retaining the other 7), and the results are shown in Table \ref{tab:subfeature_ablation}.

\begin{table*}[htbp]
\centering
\footnotesize
\caption{Ablation Experiment Results of Sub-features (AUC Scores).}
\label{tab:subfeature_ablation}
\setlength{\tabcolsep}{5pt}
\renewcommand{\arraystretch}{1.2}
\begin{tabular}{l cccccccccc}
\toprule
\multirow{2}{*}{Dataset} & \multicolumn{8}{c}{Sub-feature (Ablation Variant)} & \multirow{2}{*}{Full} \\
\cmidrule(lr){2-9}
& -Abs\_Mean & -Std & -10p\_Ratio & -Sparsity & -Row\_Ecc & -Col\_Ecc & -Row\_Max & -Row\_Std \\
\midrule
WikiMIA & 0.95/0.71 & 0.95/0.72 & 0.95/0.70 & 0.94/0.67 & 0.95/0.70 & 0.93/0.62 & 0.95/0.73 & 0.95/0.72 & \textbf{0.96}/\textbf{0.84} \\
ArxivTection & 0.96/0.82 & 0.96/0.81 & 0.96/0.83 & 0.96/0.82 & 0.96/0.83 & 0.95/0.78 & 0.96/0.83 & 0.96/0.83 & \textbf{0.97/0.85} \\
\bottomrule
\end{tabular}
\end{table*}

The sub-feature ablation results verify that all sub-features contribute to detection performance, with their combination achieving optimal AUC and TPR scores across both datasets. Among these sub-features, Col\_Ecc (a Position Offset Feature) is the most critical: its removal leads to the sharpest performance decline (AUC drops to 0.93 on WikiMIA and 0.95 on ArxivTection, with TPR falling to 0.62 and 0.78 respectively). In contrast, removing Abs\_Mean, Std, 10p\_Ratio, Row\_Ecc, Row\_Max, or Row\_Std induces only mild performance degradation, with AUC remaining above 0.95 on both datasets, indicating their relatively moderate importance; Sparsity also shows a slight impact on WikiMIA, with its ablation resulting in an AUC of 0.94. Overall, the fine-grained sub-feature design is rational, as each component provides complementary discriminative information for pre-training data detection.

\subsection{Sub-module Ablation Experiment}
The ATT and FFN modules are decomposed into 7 sub-modules based on Transformer architecture details:
\begin{itemize}
    \item ATT-related sub-modules: Q-Proj, K-Proj, V-Proj, Out-Proj
    \item FFN-related sub-modules: Gate-Proj, Up-Proj, Down-Proj
\end{itemize}

Ablation is performed by removing one sub-module’s features at a time, and the results are shown in Table \ref{tab:submodule_ablation}.

\begin{table*}[htbp]
\centering
\footnotesize
\caption{Ablation Experiment Results of Sub-modules (AUC / TPR@5\% FPR Scores).}
\label{tab:submodule_ablation}
\setlength{\tabcolsep}{4pt}
\renewcommand{\arraystretch}{1.2}
\begin{tabular}{l cccccccc}
\toprule
\multirow{2}{*}{Dataset} & \multicolumn{7}{c}{Sub-module (Ablation Variant)} & \multirow{2}{*}{Full} \\
\cmidrule(lr){2-8}
& -Q-Proj & -K-Proj & -V-Proj & -Out-Proj & -Gate-Proj & -Up-Proj & -Down-Proj \\
\midrule
WikiMIA & 0.95/0.72 & 0.95/0.71 & 0.94/0.69 & 0.94/0.67 & 0.95/0.69 & 0.94/0.65 & 0.94/0.61 & \textbf{0.96/0.84} \\
ArxivTection & 0.96/0.80 & 0.96/0.82 & 0.95/0.79 & 0.96/0.80 & 0.96/0.82 & 0.95/0.78 & 0.96/0.82 & \textbf{0.97/0.85} \\
\bottomrule
\end{tabular}
\end{table*}

The sub-module ablation results demonstrate that all sub-modules provide complementary gradient information, as single-sub-module ablation consistently leads to performance degradation while integrating all sub-modules' features achieves the optimal AUC and TPR@5\% FPR scores across both datasets. Though the performance differences between individual ablation variants are mild, ATT-related sub-modules show slightly stronger discriminative power than FFN-related ones, with K-Proj performing prominently among ATT sub-modules; in contrast, among FFN sub-modules, Down-Proj and Gate-Proj contribute more to detection performance than Up-Proj, as retaining the former two yields slightly higher AUC and TPR scores. The performance gap between single-sub-module variants and the full model further confirms the necessity of multi-sub-module feature fusion, which enables the capture of comprehensive gradient patterns across the Transformer architecture. It can be seen from this that our method also achieves excellent performance with smaller LoRA adaptation parameters and lower computational complexity.

\section{Additional Experiments Analysis}
\label{app:analysis_sub}
\subsection{Low-Data Scenarios}
\label{app:low_data}
To thoroughly verify the data efficiency and robustness of our GDS method in low-data scenarios with limited labeled samples, we conduct extensive experiments where only 10\% of the training data is used for model training. Specifically, we evaluate GDS across five different language models and two benchmark datasets, with Mink++ as the baseline for performance comparison. The experimental results, measured by two key metrics (AUC score and TPR@5\% FPR), are presented in Table~\ref{tab:low_data}, which clearly demonstrates that GDS still maintains superior detection performance compared to the baseline even under the strict constraint of limited training data.

\begin{table}[htbp]
\centering
\footnotesize
\caption{Detection performance under low-data scenario (10\% training data).}
\label{tab:low_data}
\setlength{\tabcolsep}{3.5pt}
\renewcommand{\arraystretch}{1.2}
\begin{tabular}{lcccc}
\toprule
\multirow{2}{*}{Model} & \multicolumn{2}{c}{WikiMIA} & \multicolumn{2}{c}{ArxivTection} \\
\cmidrule(lr){2-3} \cmidrule(lr){4-5}
& GDS & mink++ & GDS & mink++ \\
\midrule
llama          & 0.90/0.52 & 0.82/0.22 & 0.91/0.65 & 0.83/0.22 \\
pythia         & 0.81/0.33 & 0.70/0.18 & 0.88/0.56 & 0.70/0.17 \\
opt            & 0.86/0.37 & 0.65/0.11 & 0.88/0.47 & 0.65/0.10 \\
gpt-j       & 0.84/0.37 & 0.69/0.19 & 0.91/0.60 & 0.69/0.19 \\
gpt-neo   & 0.81/0.35 & 0.67/0.15 & 0.92/0.67 & 0.67/0.15 \\
\bottomrule
\end{tabular}
\end{table}

\subsection{Sensitivity Analysis}
\label{app:sensitivity}
We conduct sensitivity experiments on LLaMA-7B (WikiMIA) by varying LoRA rank, initialization seed, and target module under main experiment settings.Table~\ref{tab:sensitivity_analysis} shows our method maintains stable performance: AUC remains high across ranks and seeds; it maintains excellent performance even when only a small number of target modules are used.

\begin{table}[!t]
\centering
\small
\caption{Sensitivity analysis on WikiMIA (LLaMA-7B) using AUROC and TPR@5\%FPR.}
\label{tab:sensitivity_analysis}
\setlength{\tabcolsep}{6pt}
\renewcommand{\arraystretch}{1.1}
\begin{tabular}{lcc}
\toprule
LoRA rank $r$ & AUC & TPR@5\%FPR \\
\midrule
16 & 0.96 & 0.84 \\
8  & 0.96 & 0.82 \\
4  & 0.95 & 0.77 \\
\midrule
Seed & AUC & TPR@5\%FPR \\
\midrule
99 & 0.9658 & 0.8421 \\
88 & 0.9674 & 0.8408 \\
77 & 0.9636 & 0.8308 \\
66 & 0.9623 & 0.8358 \\
\midrule
Target Module & AUC & TPR@5\%FPR \\
\midrule
all        & 0.96 & 0.84 \\
mlp        & 0.94 & 0.72 \\
att        & 0.95 & 0.79 \\
\bottomrule
\end{tabular}
\end{table}

\begin{table*}[!t]
\centering
\footnotesize
\renewcommand{\arraystretch}{1} 
\caption{Performance Comparison between LoRA and Full-Parameter Update using AUROC on WikiMIA and ArxivTection.}
\label{tab:full_update_comparison}
\makebox[\textwidth][c]{
\begin{tabular}{p{1.5cm} p{2cm} p{0.7cm} c p{0.7cm} c p{0.7cm} c p{0.7cm} c p{0.7cm} c}
\toprule
\multirow{2}{*}{Dataset} & \multirow{2}{*}{Metric} & \multicolumn{2}{c}{LLaMA-7B} & \multicolumn{2}{c}{GPT-J-6B} & \multicolumn{2}{c}{OPT-6.7B} & \multicolumn{2}{c}{Pythia-6.9B} & \multicolumn{2}{c}{Neo-2.7B} \\
\cmidrule(lr){3-4} \cmidrule(lr){5-6} \cmidrule(lr){7-8} \cmidrule(lr){9-10} \cmidrule(lr){11-12}
& & LoRA & Full & LoRA & Full & LoRA & Full & LoRA & Full & LoRA & Full \\
\midrule
\multirow{2}{*}{WikiMIA} & AUC & \textbf{0.96} & \textbf{0.96} & \textbf{0.93} & 0.90 & \textbf{0.94} & 0.92 & \textbf{0.92} & 0.89 & \textbf{0.90} & 0.88 \\
& TPR@5\% FPR & \textbf{0.84} & 0.79 & \textbf{0.66} & 0.58 & \textbf{0.67} & 0.56 & \textbf{0.63} & 0.50 & \textbf{0.60} & 0.51 \\
\midrule
\multirow{2}{*}{ArxivTection} & AUC & \textbf{0.97} & 0.96 & \textbf{0.97} & 0.96 & \textbf{0.94} & 0.93 & \textbf{0.95} & 0.94 & \textbf{0.94} & \textbf{0.94} \\
& TPR@5\% FPR & \textbf{0.85} & 0.84 & \textbf{0.86} & 0.79 & \textbf{0.75} & 0.67 & \textbf{0.83} & 0.73 & 0.73 & \textbf{0.75} \\
\bottomrule
\end{tabular}
}
\end{table*}

\subsection{Full Parameter Training}
\label{app:full_ft}
In LoRA training, only FFN and MLP modules are updated, while LayerNorm, despite containing meaningful training dynamics, is excluded. To account for this, we evaluate our method under full-parameter training on two small-scale datasets, BookMIA and ArxivTection, for computational feasibility. As shown in Table~\ref{tab:full_update_comparison}, although full-parameter updates cause a slight performance drop compared to LoRA, our method still outperforms most baselines. We attribute this gap to reduced gradient magnitudes in full training, which weaken discriminative signals between sample types.

\subsection{WikiMIA Modification}
\label{subsubsec:wikimia_modification_analysis}
Dataset partitioning can introduce distribution shifts or dataset-specific artifacts. For example, WikiMIA is split by publication time, with all non-member samples labeled as 2023. To ensure our method does not exploit such artifacts, we remove all timestamp tokens and reevaluate all methods. As shown in Table~\ref{tab:wikimia_modification_analysis}, although token removal degrades performance, our method suffers a much smaller drop than the strongest baseline, FSD, and still achieves SOTA results.

\begin{table}[!t]
\centering
\small
\caption{Comparison of Temporal Feature Perturbation via AUROC. Deletion denotes the removal of all year timestamps.}
\label{tab:wikimia_modification_analysis}
\setlength{\tabcolsep}{10pt} 
\begin{tabular}{cccccc}
\toprule
Method & Origin & Deletion \\
\midrule
PPL          & 0.69 &  0.62  \\
Min-k        & 0.73 &  0.63  \\
Min-k++      & 0.82 &  0.59  \\
FSD          & 0.92 & 0.76  \\
Ours         & \textbf{0.96} & \textbf{0.84} \\
\bottomrule
\end{tabular}
\end{table}

\end{document}